\documentclass[11pt,a4paper]{article}

\newcommand{\PathOPENTitle}{PathOPEN: A pathologist-curated histopathology dataset for visual question answering}
\newcommand{\PathOPENResourceLink}[3]{%
  \raisebox{-0.25\height}{#2}%
  \enspace\textbf{#3:}\enspace
  \href{#1}{\textcolor[HTML]{214B6C}{\nolinkurl{#1}}}%
}
\newcommand{\PathOPENResourceLinks}{%
  \vspace{-0.65em}%
  {\centering\small
  \PathOPENResourceLink{https://github.com/embedded-robotics/path-open-data}{\includegraphics[height=1.1em]{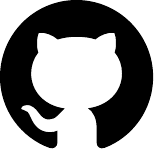}}{Code}\\[2pt]
  \PathOPENResourceLink{https://doi.org/10.5281/zenodo.24358341}{\includegraphics[height=1.1em]{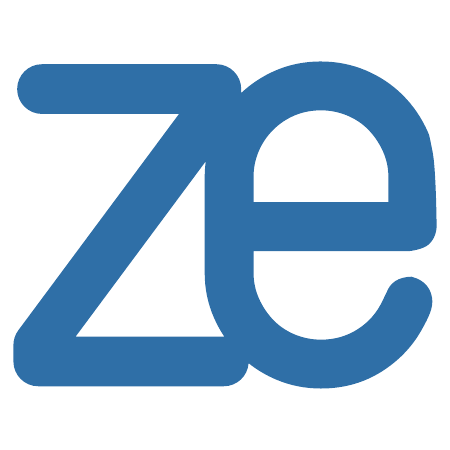}}{Dataset}\\[2pt]
  \PathOPENResourceLink{https://huggingface.co/datasets/xuzhl/PathOPEN}{\includegraphics[height=1.1em]{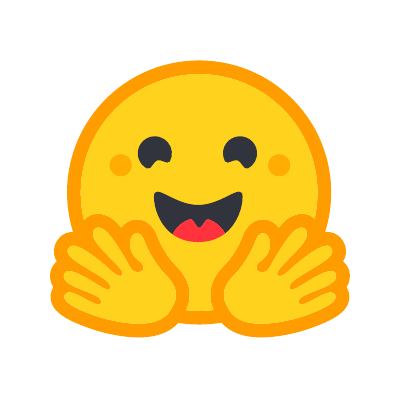}}{Hugging Face}%
  \par}%
}

\usepackage[T1]{fontenc}
\usepackage[utf8]{inputenc}
\usepackage{lmodern}
\usepackage[a4paper,margin=1in]{geometry}
\usepackage{microtype}
\usepackage{graphicx}
\usepackage{amsmath,amssymb,amsfonts,amsthm,mathrsfs}
\usepackage{textcomp,xcolor,booktabs,multirow,tabularx,array}
\usepackage[title]{appendix}
\usepackage{float,enumitem,changepage}
\usepackage{caption,subcaption}
\usepackage{siunitx}
\usepackage{authblk}
\usepackage[numbers,sort&compress,square]{natbib}
\usepackage[hidelinks]{hyperref}
\usepackage[section]{placeins}
\newsavebox{\pathopentablebox}

\newsavebox{\pathopencaptionbox}
\newlength{\pathopenfigureheight}
\newcommand{\PathOPENFigure}[3]{%
  \begin{figure}[p]
  \centering
  \sbox{\pathopencaptionbox}{\begin{minipage}{\linewidth}\caption{#2}\label{#3}\end{minipage}}%
  \setlength{\pathopenfigureheight}{\dimexpr\textheight-\ht\pathopencaptionbox-\dp\pathopencaptionbox-2\baselineskip\relax}%
  #1\par
  \usebox{\pathopencaptionbox}
  \end{figure}
}
\providecommand{\PathOPENTitle}{A histopathology dataset curated by pathologists for visual question answering}
\title{\PathOPENTitle}
\hypersetup{pdftitle={\PathOPENTitle}}
\author[1,*]{\mbox{Awais Naeem}}
\author[2,*]{\mbox{Zhongling Xu}}
\author[3]{\mbox{George Huang}}
\author[4]{\mbox{Amy Coffey}}
\author[4]{\mbox{Nkechi Okonkwo}}
\author[4]{\mbox{Mario Saab-Chalhoub}}
\author[5]{\mbox{John Lee Sang}}
\author[5]{\mbox{Dina Kokh}}
\author[5]{\mbox{Muhammad Zulfiqar}}
\author[6]{\mbox{Ankita Shukla}}
\author[2]{\mbox{Hairong Wang}}
\author[4]{\mbox{Chandra Krishnan}}
\author[1]{\mbox{Ying Ding}}
\affil[1]{School of Information, The University of Texas at Austin, USA}
\affil[2]{Cockrell School of Engineering, The University of Texas at Austin, USA}
\affil[3]{Dell Medical School, The University of Texas at Austin, USA}
\affil[4]{Department of Diagnostic Medicine, Dell Medical School, The University of Texas at Austin, USA}
\affil[5]{Oculus Pathology, USA}
\affil[6]{Department of Computer Science and Engineering, University of Nevada, Reno, USA}

\date{}
\begin{document}
\maketitle
\begingroup
\renewcommand{\thefootnote}{\fnsymbol{footnote}}
\footnotetext[1]{Equal contribution}
\footnotetext[0]{\raggedright Correspondence to
  \mbox{Hairong Wang (\href{mailto:hairong.wang@austin.utexas.edu}{hairong.wang@austin.utexas.edu})},
  \mbox{Chandra Krishnan (\href{mailto:c.krishnan@thinkoculus.com}{c.krishnan@thinkoculus.com})},
  and \mbox{Ying Ding (\href{mailto:ying.ding@ischool.utexas.edu}{ying.ding@ischool.utexas.edu})}.}
\endgroup
\ifdefined\PathOPENResourceLinks
  \PathOPENResourceLinks
\fi
\begin{abstract}
PathOPEN is a histopathology dataset for visual question answering. Two clinical
pathologists selected 157 cases with 229 available original images and wrote
open-ended, five-option multiple-choice and close-ended (yes/no) questions. They supplied both a
correct and a wrong answer for 201 of the 309 open-ended questions. Each case records
the organ, pathology category and regional anatomy, and each image records its
magnification. The dataset includes 6,870 augmented images and four manifests
that select subsets to improve balance by individual metadata field. Five
additional pathologists and two vision-language model judges rated items with
scoring rubrics, and six models answered questions zero-shot across the three
formats. The metadata support subgroup evaluation, and the paired answers can
be used to construct preference examples.
\end{abstract}
\medskip
\section{Background \& Summary}\label{sec:background}

Histopathology visual question answering (VQA) tests whether a model can interpret
a tissue image in the context of a question. An explanatory reference answer describes
the findings that support a diagnosis. A paired wrong answer shows how a plausible
interpretation can fail. Recording the organ and magnification also lets users
compare performance across image groups.

PathVQA~\cite{he2020pathvqa30000questionsmedical} draws on textbook figures and
captions, and Quilt-VQA~\cite{seyfioglu2023quiltllava} pairs images from educational
histopathology videos with human-authored evaluation questions. Their open-ended
records in our comparison provide reference answers without paired wrong answers.
PathMMU~\cite{sun2024pathmmumassivemultimodalexpertlevel} and
PatchVQA~\cite{ma2025pathbenchcomprehensivecomparisonbenchmark} provide model-assisted
multiple-choice items with expert validation. Whole-slide
benchmarks~\cite{chen2024wsivqainterpretingslideimages,chen2025slidechatlargevisionlanguageassistant}
and pathologist-authored datasets~\cite{buckley2026navigatinggigapixelpathologyimages}
support slide-level interpretation and expert-authored questions. The comparison
records used in this study lack the combination of organ-system and per-image
magnification annotations used in our subgroup analyses. Open-ended reference
length also varies across the datasets (Section~\ref{sec:pillar4}).

Two clinical pathologists selected 157 hospital and teaching cases for PathOPEN,
with 229 available original images. For each case, they wrote one or two open-ended
questions, one five-option multiple-choice question (MCQ) and one close-ended
(yes/no) question.
Of the 309 open-ended questions, 201 have both a correct and a wrong answer. The
instructions called for questions that require the image, answers that explain the
reasoning and MCQ options that can be read as complete clinical statements
(Section~\ref{sec:methods}). Each available image has 30 augmentations. Four CSV
manifests select image subsets to improve balance by organ, categorization, regional
anatomy or magnification (Section~\ref{sec:data-records}).

Pathologists gave PathOPEN higher answer-quality and grounding scores than
filtered PathVQA in the rated comparisons. The two vision-language model (VLM)
judges showed low agreement with pathologists on wrong-answer tiers and ranked
MCQ distractors differently. Technical Validation also reports ratings on the
augmentation evaluation set, changes in class balance and zero-shot performance
across question formats (Section~\ref{sec:technical-validation}).
Section~\ref{sec:usage-notes} explains how to load records, split the data by case
and use the paired answers for preference learning.

\FloatBarrier
\section{Methods}\label{sec:methods}

Figure~\ref{fig:pathopen_main_workflow} shows how two clinical pathologists
assembled the case and question records and how we generated the augmentations.
Five additional pathologists rated PathOPEN and filtered PathVQA. Two VLM judges
rated both datasets and also scored Quilt-VQA, PathMMU and PatchVQA under the
applicable rubrics (Table~\ref{tab:evaluation-design}). In a separate evaluation,
six models answered the questions zero-shot.

\PathOPENFigure{\includegraphics[width=\linewidth,height=\pathopenfigureheight,keepaspectratio]{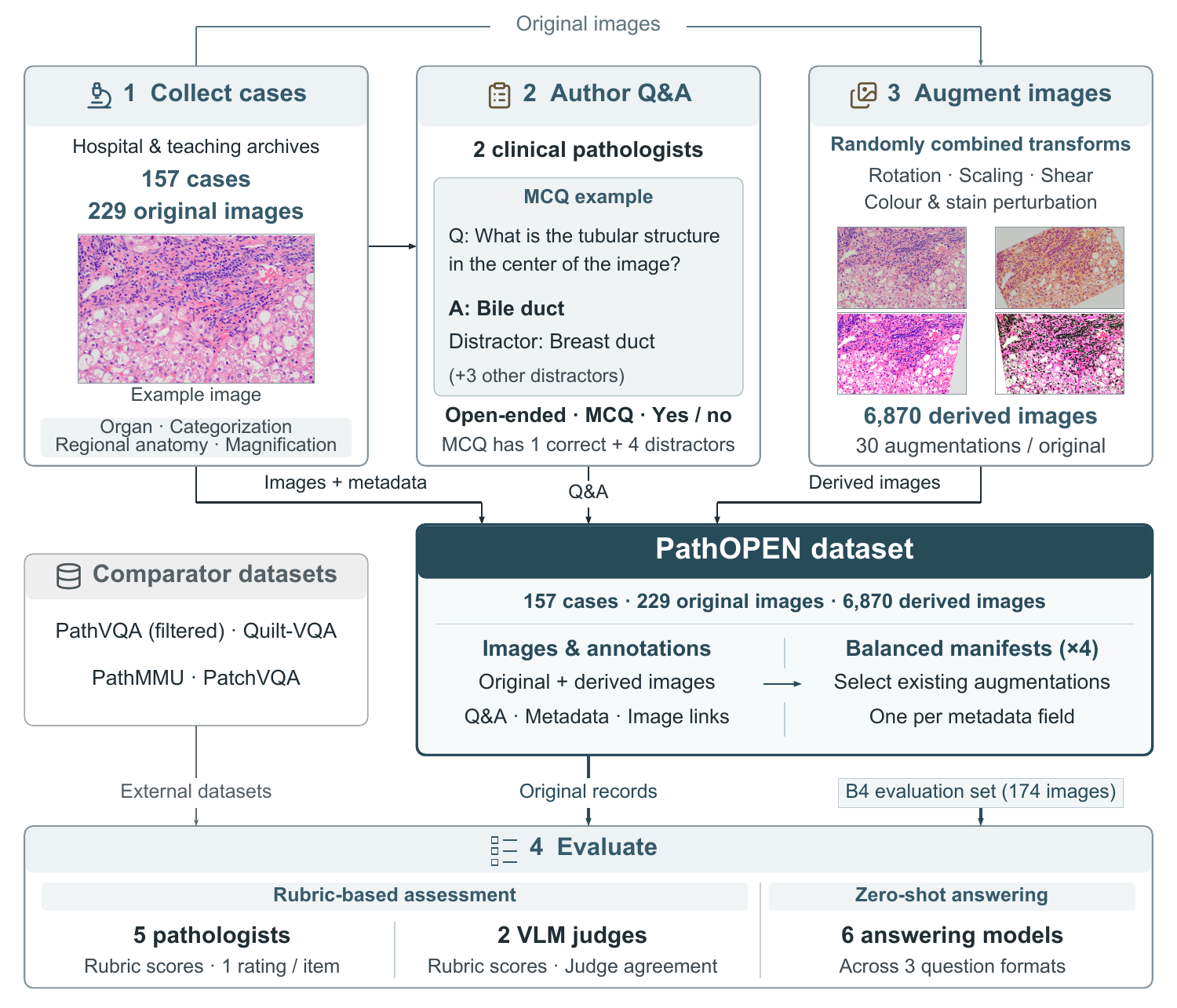}}{\textbf{PathOPEN construction and evaluation.}
Two clinical pathologists collected 157 cases and authored open-ended, multiple-choice
(MCQ) and yes/no questions. The example shows one original image, its MCQ with the
correct answer and one of four distractors, and four augmentations of the same image.
The listed transformations are examples from the augmentation pool. Applying 30
augmentations per available original image yielded 6,870 derived images from 229 originals.
Derived images inherit their source questions, answers and metadata. The PathOPEN
dataset includes these records and four balanced manifests that select existing
augmentations separately for each metadata field. Selections can overlap.
Original records and comparator datasets enter the relevant evaluations.
Benchmark~4 uses a separate evaluation set of 174 augmented images
(Section~\ref{subsubsec4_5_3}). Five additional pathologists and two
vision-language model (VLM) judges score items using rubrics.
Six answering models respond zero-shot across the three question formats.
Table~\ref{tab:evaluation-design} gives the dataset and rater scope for each rubric benchmark.}{fig:pathopen_main_workflow}

\subsection{Case collection and image acquisition}\label{subsec4_1}

Over three months, the pathologists selected 157 adult and paediatric cases from
hospital material and an internal teaching archive at Dell Medical School,
UT Austin. The pathologists favoured teaching cases marked as
interesting and hospital cases encountered during the collection period. They
selected findings that would challenge a pathology resident and sought a balance
between normal and abnormal cases, including tumours. The images come from
high-resolution whole-slide images (WSIs).

A microscope-based WSI scanner provided a maximum objective resolution of 0.147~$\mu$m per pixel
at 40$\times$ objective magnification. Digital patches were captured with a 2.3-megapixel
still-image sensor (1920~$\times$~1200 pixels) connected to the microscope through a
10$\times$ C-mount camera adapter. The pixel dimensions of the released original images vary.
The archived teaching WSIs carried no patient-sensitive information and required
no additional de-identification. For hospital cases, identifiers were removed, including
names, dates, addresses, email addresses, social security numbers, and device or biometric
identifiers. The study's IRB determination is described in Section~\ref{subsec:ethics}.

The pathologists entered metadata, questions and answers in a Google Form with
dropdown and free-text fields. In an initial trial, questions written at different
magnifications of the same WSI differed little. The final protocol allowed patches
at up to five magnifications per case and required each question and answer to apply
to every image in the case. The multi-image cases in the collection have three
image slots (Table~\ref{tab:data-records}). The pathologists recorded organ,
categorization and regional anatomy for each case and magnification for each image.
One pathologist recorded objective magnification alone. We multiplied those entries
by ten to express both sets of readings as objective magnification times the
10$\times$ eyepiece factor (Section~\ref{subsec:metadata-records}).

\subsection{Question--answer authoring}\label{subsec4_2}

Each case could contain up to two open-ended questions, one five-option
multiple-choice question (MCQ) and one close-ended question. Pathologists were instructed to make all three
formats relevant to the supplied images.

\paragraph{Open-ended questions.}
Each question required the image and a correct answer. The instructions specified
at least 15--20 words for the correct answer to include diagnostic reasoning and
context. The same pathologist could also write a plausible wrong answer. They did so
by introducing a subtle factual, theoretical or hypothetical error, or by writing
an answer that would be appropriate for a different diagnosis. These matched
answers can be used to construct preference examples.
Section~\ref{subsec:qa-records} reports the counts.

\paragraph{Multiple-choice questions.}
MCQs were designed to support both option selection and conversion into open-ended
question--answer records. The correct option supplies the correct answer; each of the
four distractors supplies a wrong answer to the same question. Stems therefore used
open-ended phrasing and omitted selection prompts such as \emph{Select from the following
options}, \emph{Which of the following}, \emph{Select the best one}, \emph{Out of the following
options} and \emph{Select 1 out of 5}. The instructions asked for four to six words per option, five options relevant
to the image and one correct choice. Distractors were intended to differ subtly
from the correct option. Many options exceeded the requested length
(Supplementary Note~1). Benchmark~5 assesses each option on its own
(Section~\ref{subsubsec4_4_5}).

\paragraph{Close-ended questions.}
Close-ended questions required a \emph{Yes} or \emph{No} answer without further explanation,
while retaining contextual relevance to the image. Typical openings were \emph{Is},
\emph{Are}, \emph{Can} and \emph{Could}. This format also enabled comparison with PathVQA,
which includes both open-ended and close-ended items.

\subsection{Image augmentation and balanced-manifest construction}\label{subsec4_3}

\subsubsection{Image generation}\label{subsubsec:augmentation-generation}
Each available original image was augmented 30 times. For each augmentation, 5 of 16
transforms were sampled, assigned magnitudes drawn from 0--5, and applied successively
in random order. The transform library described in \cite{FARYNA2024108018} included
Scaling, TranslateX, TranslateY, ShearX, ShearY, Brightness, Sharpness, Contrast, Rotate,
Equalize, Hue, Saturation, Value, Hematoxylin, Eosin and DAB. After consultation with the
pathologists, elastic deformation, Gaussian blur and Gaussian noise were omitted because
high magnitudes could make the original structures unidentifiable. The source study tested the transforms for model generalisation on haematoxylin
and eosin (H\&E) images. We used Benchmark~4 to assess diagnostic content in PathOPEN's
augmented images (Section~\ref{sec:pillar5}).

\subsubsection{Class-balanced variants}\label{subsubsec:balancing}
Uniform augmentation multiplies every class count by the same factor and leaves class
proportions, Gini coefficients and entropy unchanged. Balanced variants therefore retain
different numbers of augmentations from different source images.

For each metadata field, the number retained per image is
\[\min\!\left(n_{\mathrm{available}},
  \operatorname{round}\!\left[\mathrm{clip}(n_{\max}/n_{\mathrm{class}}, 1, 10)\right]\right),\]
where $n_{\mathrm{class}}$ is the number of images in that image's class, $n_{\max}$ is
the largest class count, and $n_{\mathrm{available}}$ is the number of available
augmentations (30 per released image). The clipped multiplier is rounded to an integer
using pandas \texttt{round}. A seeded shuffle (seed 20260817) selects the augmentation
files. An image in the largest class contributes one augmentation. Images in rare
classes can contribute up to ten. The cap limits the number of versions of any one
source image. Section~\ref{sec:pillar5} shows how varying the cap changes class balance.

Organ, categorization, regional anatomy and magnification are balanced separately.
Each output is a manifest selecting existing files, so constructing a variant generates
no additional images and leaves all 30 augmentations per source image in the full
release. Balance is measured by the largest-to-smallest class ratio, Gini coefficient
and normalised Shannon entropy. Section~\ref{subsec:file-records} describes the file
structure, and Section~\ref{subsec:augmentation-records} gives the four manifest sizes.

\subsection{Comparator dataset selection}\label{subsec:pathvqa-selection}

\paragraph{PathVQA.}
PathVQA includes images other than H\&E-stained histopathology. We used
HistoCartography~\cite{jaume2021histocartographytoolkitgraphanalytics} to detect nuclei and
retained images with more than 10 detected nuclei as a proxy for H\&E content. Of the
4,998 images and 32,795 question--answer pairs in the released files, this retained
1,241 images and 8,992 pairs. The PathVQA publication reports 32,799 pairs; counts here
refer to the released files to which the filter was applied.

Answers of \emph{yes} or \emph{no} defined the close-ended subset; all other answers defined
the open-ended subset. The image-filtered set contained 4,630 close-ended and 4,362
open-ended pairs. We then required at least five words in both the question and answer for
open-ended pairs and in the question alone for close-ended pairs. This left 612
open-ended and 3,005 close-ended pairs, about 11\% of the original corpus
(Figure~\ref{fig:pathvqa-selection-text}a). The length filter excluded 86\% of the
open-ended pairs that passed image selection. The evaluation sheets used the first
50 eligible close-ended pairs in the concatenated train, test and validation split
order, before shuffling and allocation to pathologists. The six answering models
were evaluated on 612 open-ended and 50 close-ended items. Section~\ref{subsubsec4_5_2}
reports the historical human rating counts. The word-count summaries in
Supplementary Note~1 use the image-filtered set before the length filter.

\paragraph{Quilt-VQA, PathMMU and PatchVQA.}
We used the \texttt{train} split of \texttt{wisdomik/Quilt\_VQA}. For closed
questions, the loader retained records with an identifiable yes or no verdict
and used that verdict as the reference answer. It loaded the \texttt{test\_tiny}
splits for PathMMU and PatchVQA, retained resolvable images and excluded the
shared \texttt{PathCLS} category from PatchVQA to avoid counting that source
twice. The zero-shot comparison used 724 open-ended and 257 close-ended
Quilt-VQA items, 905 PathMMU MCQs and 294 PatchVQA MCQs
(Table~\ref{tab:dataset-comparison}). Table~\ref{tab:evaluation-design} gives the
scope of each rubric assessment.

\subsection{Evaluation benchmarks}\label{subsec4_4}

Five benchmarks assess answer quality, image grounding and augmentation quality.
Table~\ref{tab:evaluation-design} identifies their rated units, datasets and rater sources.
Raters scored evaluable items from 0 to 2 and used $-1$ when the question or
image could not be understood or did not support a judgement. We report the frequency
of $-1$ separately and specify how each analysis handles it
(Section~\ref{subsubsec:stats}).

\begin{table}[tbp]
\centering\small
\caption{\textbf{Evaluation design.} Each row identifies the object receiving a rubric
score. MCQ-derived correct and wrong answers are rated under Benchmarks~1 and 2 as
open-ended records; Benchmark~5 separately assesses the options in their MCQ role.
The full rubrics are in Supplementary Tables~\ref{tab:benchmark1}--\ref{tab:benchmark5b}.}
\label{tab:evaluation-design}
\begin{tabularx}{\linewidth}{@{}l>{\raggedright\arraybackslash}X>{\raggedright\arraybackslash}X>{\raggedright\arraybackslash}X@{}}
\toprule
Benchmark & Rated unit & Pathologist-rated datasets & VLM-judge datasets \\
\midrule
1 & Question, image and correct answer & PathOPEN; filtered PathVQA &
PathOPEN; filtered PathVQA; Quilt-VQA \\
2 & Question, image and wrong answer & PathOPEN & PathOPEN \\
3 & Close-ended question, image and answer & PathOPEN; filtered PathVQA &
PathOPEN; filtered PathVQA; Quilt-VQA \\
4 & Augmented image with an open-ended question and correct answer & PathOPEN augmentations &
PathOPEN augmentations \\
5a/5b & MCQ correct option / individual distractor, with question and image & None &
PathOPEN; PathMMU; PatchVQA \\
\bottomrule
\end{tabularx}
\end{table}

\subsubsection{Benchmark 1: correct open-ended answers}\label{subsubsec4_4_1}
Knowledge Interpretation and Deduction measures whether an answer addresses the main question
with appropriate reasoning and without unrelated information. Visual Grounding measures
whether it uses relevant image information to answer the question
(Supplementary Table~\ref{tab:benchmark1}).

\subsubsection{Benchmark 2: wrong open-ended answers}\label{subsubsec4_4_2}
Error Proximity and Deductive Plausibility measures conceptual and reasoning proximity
to the correct answer. Visual Grounding Error measures whether the wrong answer attends
to the relevant image region and features. Higher scores mark subtler errors in reasoning or visual interpretation
(Supplementary Table~\ref{tab:benchmark2}).

\subsubsection{Benchmark 3: close-ended answers}\label{subsubsec4_4_3}
Visual Grounding/Reasoning measures whether answering \emph{Yes} or \emph{No} requires
combining question text with image evidence, with higher scores assigned when both
are needed (Supplementary Table~\ref{tab:benchmark3}).

\subsubsection{Benchmark 4: augmented images}\label{subsubsec4_4_4}
Clinical Relevance measures whether an augmented image remains usable for diagnosis.
Visual Grounding measures whether its visual content, together with the question,
supports the supplied answer (Supplementary Table~\ref{tab:benchmark4}).

\subsubsection{Benchmark 5: MCQ standalone assessment and grounding}\label{subsubsec4_4_5}
Benchmark~5a rates the correct option for Standalone Completeness and Visual Grounding;
Benchmark~5b rates each distractor separately for Distractor Standalone Plausibility and Distractor
Visual Grounding Error (Supplementary Tables~\ref{tab:benchmark5a} and \ref{tab:benchmark5b}).
The rubrics assess whether an option is a complete clinical statement and
whether judging it requires image evidence.

\subsection{Pathologist rating protocol}\label{subsec4_5}

Five volunteer clinical pathologists were recruited to evaluate PathOPEN, filtered
PathVQA and augmented images. Each sheet presented an image, its question and answer,
and the relevant criteria, with no column identifying the source dataset. Each worksheet row was assigned to one
pathologist.

\subsubsection{PathOPEN items}\label{subsubsec4_5_1}
Four pathologists each received 46 image-level evaluation rows, and the fifth received 47.
A row contained one image and its case's 1--2 open-ended questions, 1 MCQ and 1
close-ended question; images from a multi-image case occupied separate rows.
Correct open-ended answers were rated under Benchmark~1 and available wrong answers
under Benchmark~2. MCQs were converted to open-ended records by presenting the correct
option as a correct answer and the four distractors as wrong answers, scored under
Benchmarks~1 and 2 respectively. Close-ended answers were rated under Benchmark~3.

\subsubsection{Filtered PathVQA items}\label{subsubsec4_5_2}
The historical sheets contain 612 rows with an open-ended question and answer,
each with two Benchmark~1 scores, and 50 close-ended rows with Benchmark~3
scores. The close-ended rows also contain 12 entries in the Benchmark~1 Knowledge
Interpretation and Deduction column and 10 in the Visual Grounding column.
The saved Benchmark~1 analyses pool numeric entries from these columns, giving
624 and 622 scores. These summaries describe the historical score columns.
The zero-shot comparisons use the 612 open-ended pairs identified by the
selection procedure (Section~\ref{subsec:pathvqa-selection}).

\subsubsection{Augmented images}\label{subsubsec4_5_3}
Benchmark~4 used a separate set of 174 augmented images. The evaluation
spreadsheets linked the augmented image identifiers to the source records and
presented each image with up to two original open-ended questions and correct
answers. These answers gave pathologists diagnostic details against which to assess
the augmented image. The five pathologists rated separate portions of this
set, with one human rating per item. We compared each historical image with the file of the same name stem in the
v1.0.0 dataset. All 174 pairs had matching dimensions and different decoded RGB
pixel values. The ratings refer to the historical images.

\subsubsection{MCQ standalone assessment}\label{subsubsec4_5_4}
Benchmark~5 was evaluated only by VLM judges on PathOPEN, PathMMU and PatchVQA.
At this scale, we used model judges for the additional datasets.
Section~\ref{sec:pillar3} reports each judge's scores and their agreement.

\subsection{Computational evaluation and statistical analysis}\label{subsec:computational}

The judges scored supplied answers using the rubric. The six answering models
generated their own responses. Section~\ref{sec:technical-validation} reports the
results for both tasks.

\subsubsection{Vision-language model judges}\label{subsubsec:judges}
Qwen3-VL-32B (thinking variant) and InternVL3.5-38B were used as judges. Each receives an
image, the item and the verbatim rubric, and returns a score on $\{-1,0,1,2\}$ with a
justification. Decoding is greedy with sampling disabled. The token budget is 8,192,
with one retry at 12,288 if a response is truncated before committing to a score.
The judges are larger than the 4--8B answering models.

For Benchmarks~1--4, judge scores are compared with the pathologist rating on the same
item. Agreement is reported as exact agreement, quadratic-weighted Cohen's $\kappa$
and Gwet's AC1. For Benchmark~5, agreement is computed between the two judges. We report each
judge's scores separately. A small number of Qwen distractor-criterion responses
remain truncated after retry and are recorded as $-1$. Their frequencies are reported
with the affected results.

\subsubsection{Zero-shot answering}\label{subsubsec:zeroshot}
Six models were evaluated with greedy decoding in all three question formats:
Qwen2.5-VL-7B-Instruct, InternVL3.5-8B, MedGemma-4B, LLaVA-Med v1.5 (Mistral-7B),
Quilt-LLaVA v1.5-7B and Patho-R1-7B. Two are general-purpose models and four are adapted
to medical or pathology data.

Models that emit a bare answer receive 16 new tokens. LLaVA-Med and Quilt-LLaVA receive
96 to accommodate their short preambles. Patho-R1 receives 256 because it emits a
reasoning trace before its answer. For Patho-R1, a budget sweep on 20 PathOPEN
MCQs found that parse rate and accuracy plateaued at 256: 17 of 20 responses parsed at budgets of 256, 512
and 768, compared with 12 of 20 at 16.

MCQ responses are parsed for an option letter or a matching option string.
Unparseable responses count as incorrect, so accuracy includes all items. For
cross-dataset comparisons, accuracy above each dataset's chance level accounts for
option counts ranging from two to sixteen. Close-ended responses are scored by exact
yes/no agreement. Open-ended responses are scored with token F1 and BERTScore F1.
Cross-dataset BERTScore F1 is rescaled against an empirical baseline, so zero denotes
similarity no greater than unrelated text. The two metrics give near-identical model
rankings on PathOPEN ($\rho=0.94$). We used token F1 in subgroups with few items because it
requires no baseline.

\subsubsection{Statistical analysis}\label{subsubsec:stats}
Ordinal score distributions are compared using two-sided Mann--Whitney $U$ tests, with
rank-biserial correlation as the effect size and percentile bootstrap 95\% confidence
intervals from 2,000 resamples. Comparisons addressing the same question use
Holm correction within that family. Kruskal--Wallis tests compare more than two groups,
and Spearman's $\rho$ measures rank correlation across question formats.

Dataset comparisons under Benchmarks~1 and 3, per-pathologist summaries, rater-distribution tests
and Benchmark~4 summaries retain encoded $-1$ values. Their means and rank tests
therefore reflect whether the item could be scored as well as how it was rated. Judge--pathologist agreement under Benchmarks~1--4 also retains $-1$.
Quadratic weights treat $-1,0,1,2$ as equally spaced, penalising disagreement between
$-1$ and 2 more than disagreement between 0 and 2. Benchmark~2 tier analyses exclude
$-1$, which has no difficulty tier. Blank and non-numeric entries are excluded as missing.

Each item rated by a human has one pathologist rating. We compared score
distributions across the pathologists' separate item pools. Fleiss' $\kappa$ and
Krippendorff's $\alpha$ would require overlapping ratings. Because scores are
concentrated at the maximum on PathOPEN, we report AC1 alongside $\kappa$.
The cross-dataset performance bootstrap treats each (model, item) pair as
independent. Dependence among questions or images within a case, or among model
responses to the same item, may make these intervals too narrow.

\subsection{Ethical approval declaration}\label{subsec:ethics}
On 24 July 2025, the University of Texas at Austin Institutional Review Board determined
that the activities described in submission \texttt{STUDY00007826} did not constitute
human subjects research. The submission covered case collection, evaluation by
pathologists and external sharing of de-identified pathology images, question--answer
pairs and metadata. Archived teaching material carried no patient identifiers.
Identifiers were removed from hospital material before question--answer construction
and evaluation.

The evaluating pathologists assessed the dataset as investigators. They received the
de-identified images, questions and answers, along with the scoring criteria for
Benchmarks~1--4. Benchmark~5 used VLM judges only. The determination letter is available
on request.

\FloatBarrier
\section{Data Records}\label{sec:data-records}

PathOPEN contains case records, original and augmented images, and four manifests
that select subsets of the augmented images. Table~\ref{tab:data-records} counts
the records at collection and in the dataset package. See the Data
availability statement for repository, version and licence details.

\subsection{Files, fields and record relationships}\label{subsec:file-records}

The package contains the following files and directories.
\begin{itemize}
\item \texttt{data.json} contains one object per case. Each object has a unique
\texttt{Case ID}. Case metadata are stored in \texttt{Organ}, \texttt{Categorization} and
\texttt{Regional Anatomy}. The \texttt{Images} list links to the image records.
\texttt{Open Ended} lists questions with correct answers and optional wrong answers.
\texttt{MCQ} stores a question, five options and an answer key. \texttt{Close Ended}
stores a question and its answer.
\item \texttt{images/<image\_id>/} holds each available original image and its
30 augmentations. Image identifiers follow
\texttt{img\_pathopen\_<case>\_<slot>}.
\item \texttt{manifests/} contains four \texttt{balanced\_by\_*.csv} files selecting
existing augmentation files. Each row records the selected filename and path, source
image and case, balanced field and class, magnification, number retained from that
source image's 30 augmentations, and selection rank.
\item \texttt{README.md}, \texttt{load\_pathopen.py}, \texttt{provenance.json} and
\texttt{SHA256SUMS} document the fields, provide loading and verification, and record
the source version and file checksums.
\end{itemize}

The data-processing workspace stores the original responses and collection links in
\texttt{raw/data.csv} and maps links to image identifiers in \texttt{gdrive\_mappings/}.
The released package uses image identifiers and package-relative paths in both
\texttt{Directory} and manifest \texttt{relative\_path} fields.

The \texttt{Images} list links a case to its image records. Each image record supplies
its \texttt{Directory}, \texttt{Original} filename, magnification and \texttt{Augmented}
filenames. Pair each question with every available image for its case. Expanding the
five MCQ options gives one correct-answer record and four wrong-answer records
for each image.

For example, \texttt{img\_pathopen\_148\_01} identifies image slot 01 of case 148.
The file \texttt{img\_pathopen\_148\_01\_aug\_0.png} is an augmentation of that
image. The original and its augmentations share the questions and answers for
that case. A manifest keeps the case and source image
identifiers when it selects an augmentation, so the selected file can follow its
case into the same split (Section~\ref{subsec:usage-splits}).

The original files for cases 9 and 79 could not be retrieved from the collection
store. Both cases retain their questions, metadata and an empty image slot to
match the raw responses. Their raw magnification responses are 2$\times$ and
4$\times$, respectively. Section~\ref{subsec:usage-loading} explains how to
handle these missing images and open-ended questions with only a correct answer.

\subsection{Cases, images and metadata}\label{subsec:metadata-records}

Of the 157 cases, 120 have one image slot and 37 have three. We collected 231
image slots and recovered 229 original files. The available counts in
Table~\ref{tab:data-records} use those 229 originals.

\texttt{Organ} records the organ system and subcategory
(Appendix~\ref{sec:appendix_organs}). \texttt{Categorization} records the broad
pathology category (Appendix~\ref{sec:appendix_categories}).
\texttt{Regional Anatomy} supplies a more specific cellular or anatomical description
within the organ when relevant, for example a particular region within an ovary.
These three fields describe the case. For each image, magnification records
the product of the objective magnification (2$\times$ to 40$\times$) and the
10$\times$ eyepiece factor (Section~\ref{subsec4_1}).

Metadata values retain their collected spelling and case. Axis abbreviations in
Figure~\ref{fig:composition} correspond to the full stored values listed in
Supplementary Table~\ref{tab:label-abbreviations}. Grouping labels that differ
only in capitalisation reduces regional anatomy from 113 to 111 classes and
categorization from 33 to 32.

\subsection{Questions, answers and image-pair counts}\label{subsec:qa-records}

All 309 open-ended questions have a correct answer. Of these, 201 also have a
wrong answer and 108 have only the correct answer. The 120 single-image cases contribute 235 questions,
127 with both answer types and 108 with correct answers only. The 37 three-image cases
contribute 74 questions, all with both answer types (Figure~\ref{fig:composition}e).
Pairing these questions with the available images yields 453 correct-answer and 347
wrong-answer question--image records.

Each case contains one five-option MCQ and one yes/no question. Four distractors
per image give 924 records for the 231 collected image slots and 916 for the
229 available originals. The Benchmark~2 spreadsheets contain two scores for
each of 920 distractors. They cover the two slots with missing images and omit
the four unrated distractors for case 64 (Section~\ref{sec:pillar2}).
Section~\ref{sec:usage-notes} explains the 912 records linked by image and option
position and the 910 that also match the historical answer text.

The 229 available originals also give 229 correct MCQ option records and 229
close-ended question--image pairs. Pooling native open-ended records with MCQ-derived
records gives 682 correct-answer and 1,263 wrong-answer records. Each record pairs
an answer with one image. Table~\ref{tab:data-records} gives the collected, available
and augmented totals.

\subsection{Augmented images and balanced manifests}\label{subsec:augmentation-records}

The 229 available originals each have 30 augmentations, giving 6,870 derived image files.
Each inherits its source image's questions, answers, metadata and magnification.
The transformation procedure and retention rule are specified in
Section~\ref{subsec4_3}; training benefits were not evaluated in this study.

Pairing all augmentations with their source questions multiplies the
original-image counts by 30. Table~\ref{tab:data-records} lists those totals by
answer type, including the MCQ counts when each option is treated as an answer.
Benchmark~4 assesses diagnostic content on the separate evaluation set
(Section~\ref{sec:pillar5}).

The four balanced manifests select 852 augmentation files for organ, 988 for
categorization, 427 for magnification and 994 for regional anatomy.
Rare-class images contribute more of their 30 augmentations than common-class images.
Each manifest targets one field and lists paths to existing files. Selections
can overlap. The full package contains all 6,870 derived images.

\sbox{\pathopentablebox}{%
\begin{tabular}{l S[table-format=6.0] S[table-format=6.0] S[table-format=6.0]}
\toprule
 & \multicolumn{2}{c}{Original images} & {Augmented} \\
\cmidrule(lr){2-3} \cmidrule(lr){4-4}
Item & {As collected} & {Available} & {Available} \\
\midrule
Cases & 157 & 157 & {--} \\
Images & 231 & 229 & {--} \\
Open-ended question--image pairs, correct answer & 457 & 453 & 13590 \\
Open-ended question--image pairs, wrong answer & 349 & 347 & 10410 \\
MCQ question--image pairs & 231 & 229 & 6870 \\
MCQ distractors (4 per question--image pair) & 924 & 916 & 27480 \\
Close-ended question--image pairs & 231 & 229 & 6870 \\
Pooled open-ended + MCQ, correct answer & 688 & 682 & 20460 \\
Pooled open-ended + MCQ, wrong answer & 1273 & 1263 & 37890 \\
\midrule
Augmentations per image & {--} & {--} & 30 \\
Augmented image files & {--} & {--} & 6870 \\
\bottomrule
\end{tabular}
}

\begin{tableorg}[t]
\centering
\caption{\textbf{PathOPEN record counts for collected slots and available files.}
Each question applies to every image in its case. The collected counts include the
image slots for cases 9 and 79, which retain their text and metadata. The available
counts include only records linked to files. Augmented counts are 30 times the
available original-image counts. MCQ denotes multiple-choice question. Pooled correct
and wrong counts combine native open-ended answers with individual MCQ options.
Section~\ref{sec:data-records} describes the units and relationships.}
\label{tab:data-records}
\resizebox{\linewidth}{!}{\usebox{\pathopentablebox}}
\end{tableorg}

\FloatBarrier

\section*{Data Overview}

Figure~\ref{fig:composition} summarises the case, image and question distributions.
Categorization is the most concentrated metadata field (case-level Gini 0.66),
and regional anatomy is the least concentrated (0.24). Among the cases with available
images, 118 have one recorded magnification and 37 have more than one.
\hyperref[note:metadata]{Supplementary Note~2} gives the case- and image-weighted
Gini coefficients and the combinations of magnifications within cases.
Figure~\ref{fig:pathvqa-selection-text} tracks PathVQA selection and illustrates
word lengths before and after image filtering. Supplementary Note~1 gives the
word-count rules and detailed text summaries.

\PathOPENFigure{\includegraphics[width=\linewidth,height=\pathopenfigureheight,keepaspectratio]{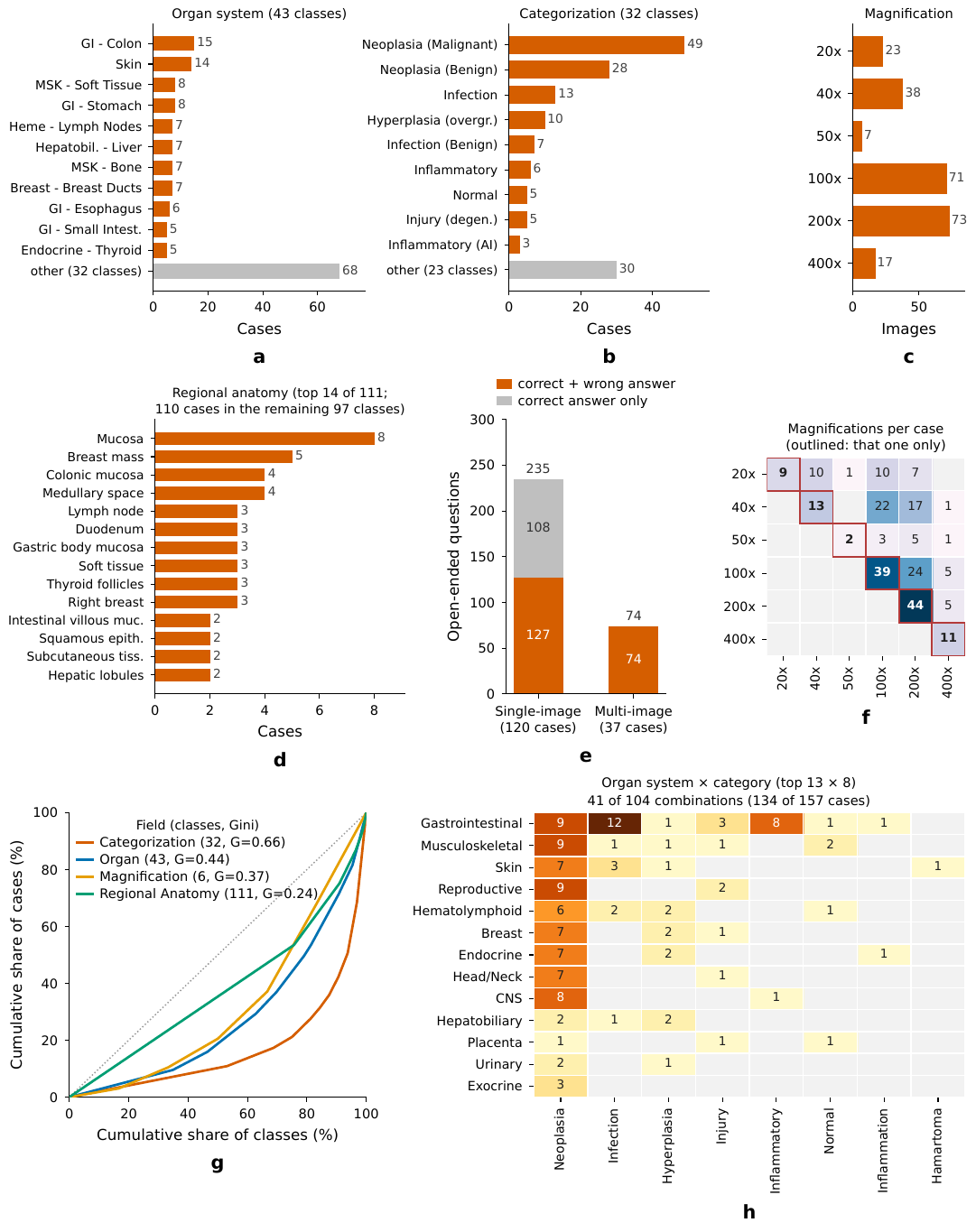}}{\textbf{Composition of the PathOPEN metadata.}
\textbf{a}, \textbf{b}, Case counts by organ system and categorization.
\textbf{c}, Counts of the 229 available original images by magnification. \textbf{d}, The fourteen most frequent
regional anatomy classes shown in the plot, with 110 cases across the remaining
97 classes. \textbf{e}, Question--answer composition by item type.
\textbf{f}, Magnification co-occurrence within cases: diagonal cells count cases with
only that magnification, and off-diagonal cells count cases with both magnifications.
\textbf{g}, Lorenz curves and Gini coefficients for all four metadata fields, including
every class; the diagonal represents equal class proportions. \textbf{h}, Case counts
for the 13 most frequent organ systems and 8 most frequent categories. The
displayed cells contain 134 of 157 cases, and 41 of the 104 combinations are
observed. Grey marks combinations with zero cases. Panel \textbf{g} uses case-level weighting for magnification
(\hyperref[note:metadata]{Supplementary Note~2}). Axis labels use abbreviated
display forms where needed; Supplementary Table~\ref{tab:label-abbreviations} maps them
to the full stored values.}{fig:composition}

\PathOPENFigure{\includegraphics[width=\linewidth,height=\pathopenfigureheight,keepaspectratio]{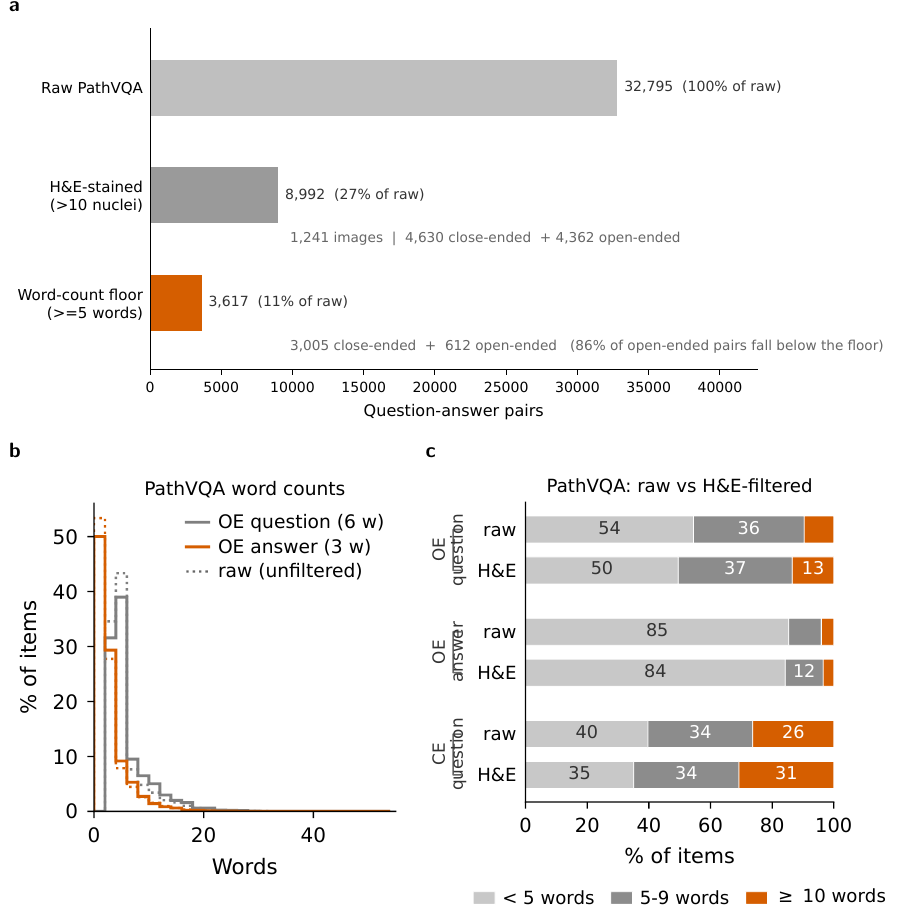}}{\textbf{Selection and text characteristics of the PathVQA comparator.}
\textbf{a}, Filtering stages. Images with more than 10 detected nuclei were retained
as a proxy for H\&E content. We then require at least five words for both the question and answer in
open-ended pairs and for the question alone in close-ended pairs. The final subset
has 612 open-ended and 3,005 close-ended pairs after applying the length filter. The length filter excludes 86\%
of open-ended pairs that passed image selection.
\textbf{b}, Open-ended question and answer length distributions for raw PathVQA
(dotted lines) and the image-filtered set (solid lines).
\textbf{c}, Percentages of open-ended questions, open-ended answers and close-ended
questions in three word-count bands ($<5$, 5--9 and $\ge 10$ words). H\&E denotes the
image-filtered set. Panels \textbf{b} and \textbf{c} compare the first two stages
in \textbf{a}, before the length filter.
Section~\ref{subsec:pathvqa-selection} gives the filtering procedure and counts.}{fig:pathvqa-selection-text}

\FloatBarrier
\section{Technical Validation}\label{sec:technical-validation}

\sbox{\pathopentablebox}{%
\setlength{\tabcolsep}{3pt}%
\begin{tabular}{l S[table-format=1.2] S[table-format=1.2] S[table-format=2.1] S[table-format=2.1] S[table-format=2.1] S[table-format=2.1] S[table-format=1.2] S[table-format=1.2] c}
\toprule
 & \multicolumn{2}{c}{Mean score} & \multicolumn{2}{c}{\% max score} & \multicolumn{2}{c}{\% unscorable} & \multicolumn{2}{c}{Rater spread} & {Effect size} \\
\cmidrule(lr){2-3} \cmidrule(lr){4-5} \cmidrule(lr){6-7} \cmidrule(lr){8-9}
Criterion & {PathOPEN} & {PathVQA} & {PathOPEN} & {PathVQA} & {PathOPEN} & {PathVQA} & {PathOPEN} & {PathVQA} & {[95\% CI]} \\
\midrule
B1 Knowledge interpretation & 1.90 & 1.14 & 95.4 & 57.2 & 0.7 & 17.9 & 0.20 & 1.01 & -0.368 [-0.415, -0.320] \\
B1 Visual grounding & 1.88 & 1.16 & 93.0 & 59.1 & 0.7 & 17.3 & 0.24 & 1.26 & -0.340 [-0.387, -0.291] \\
B3 Visual grounding & 1.96 & 0.90 & 97.0 & 58.0 & 0.0 & 34.0 & 0.09 & 2.10 & -0.399 [-0.546, -0.261] \\
\bottomrule
\end{tabular}
}

\begin{tableorg}[t]
\centering
\caption{\textbf{Answer ratings and unscorable items in PathOPEN and filtered PathVQA.}
Scores use $\{-1,0,1,2\}$; $-1$ denotes inability to evaluate. Means, effects and
per-pathologist summaries include $-1$ alongside scores for items that could be rated. Means and effects pool the five pathologists; maximum-score and unscorable
percentages average their percentages. Effects are rank-biserial correlations with
bootstrap 95\% confidence intervals (CIs), with negative values indicating higher
PathOPEN scores. Rater spread is the range of pathologist means across disjoint pools.
B1 and B3 denote Benchmarks~1 and 3. The shortened criterion labels refer to Knowledge
Interpretation and Deduction and Visual Grounding (B1), and Visual Grounding/Reasoning
(B3), as defined in Supplementary Tables~\ref{tab:benchmark1} and \ref{tab:benchmark3}.
Historical PathVQA Benchmark~1 counts include entries recorded on close-ended
rows (Section~\ref{subsubsec4_5_2}). Section~\ref{sec:pillar1} gives denominators
and tests.}
\label{tab:dataset-quality}
\sisetup{table-format=-1.3}
\resizebox{\linewidth}{!}{\usebox{\pathopentablebox}}
\end{tableorg}

\subsection{Item quality against existing pathology VQA datasets}
\label{sec:pillar1}

Five board-certified pathologists rated PathOPEN and filtered PathVQA, with one
pathologist per item. Benchmark~1 rates reasoning and image grounding for
correct open-ended answers. Benchmark~3 rates the image evidence for yes/no
answers. The mean scores and rank tests include
$-1$ ratings for items the rater could not evaluate
(Section~\ref{subsubsec:stats}). Table~\ref{tab:dataset-quality} and
Figure~\ref{fig:pillar1} report the results. Negative rank-biserial effects
indicate higher scores for PathOPEN.

\paragraph{Items that could be scored.}
Figure~\ref{fig:pillar1}b pools the recorded pathologist scores by criterion.
The first native open-ended item in each PathOPEN image slot had 3 unscorable
ratings out of 231 (1.3\%) for each Benchmark~1 criterion. In the historical
filtered PathVQA sheets, 111 of 624 Knowledge Interpretation and Deduction
ratings (17.8\%) and 107 of 622 Visual Grounding ratings (17.2\%) were
unscorable. These historical totals include scores entered in the Benchmark~1
columns on close-ended rows (Section~\ref{subsubsec4_5_2}).
Table~\ref{tab:dataset-quality} averages the five pathologists' individual rates, giving 17.9\% and 17.3\% for filtered PathVQA.
For close-ended items (Benchmark~3), pathologists marked 17 of 50 filtered
PathVQA items unscorable (34.0\%) and 0 of 230 PathOPEN items. The close-ended
score for case 64 is blank.

Each VLM judge scored 229 PathOPEN images per criterion. The filtered PathVQA
comparison used 612 open-ended items and 50 yes/no items per judge. The human
comparisons used the historical score columns described above, including PathOPEN
image slots whose originals are unavailable. The comparison with Quilt-VQA used
724 open-ended and 257 yes/no items per judge.
For the analysis across pathologists, pooling both native open-ended slots and the
MCQ-derived correct answer gives 686 PathOPEN ratings per Benchmark~1 criterion,
including 5 unscorable ratings (0.7\%).

Four of the five pathologists scored every PathOPEN item in their pool.
The largest difference in unscorable rates was 3.6 percentage points.
The widest gap on filtered PathVQA was 70 points on Benchmark~3
(Table~\ref{tab:defect-mechanism}). Each pathologist saw a different set of items,
so the rates reflect both the items assigned and the pathologist's judgements.

\paragraph{Text length and answer ratings.}
Across the 457 collected native open-ended question--image pairs, PathOPEN questions
averaged 18.7 words and correct answers 22.8, against 7.8 for each field in filtered PathVQA
(Figure~\ref{fig:pillar1}a; all two-sided Mann--Whitney $P < 10^{-4}$). Counting each
distinct question once gives PathOPEN means of 17.6 and 23.0 words, respectively
(Supplementary Note~1). The authoring protocol requested answers that explain
the reasoning. Filtered PathVQA provides no matched open-ended wrong answers for a
word-length comparison.

Pathologists gave PathOPEN higher scores on all three criteria. The
rank-biserial correlations were $-0.368$ for reasoning and $-0.340$ for image
grounding in Benchmark~1, and $-0.399$ in Benchmark~3. All three confidence
intervals exclude zero (Table~\ref{tab:dataset-quality}).
Including the two judges, eight of nine intervals exclude zero. The exception
was InternVL on Benchmark~3, with an interval from $-0.212$ to $0.008$ and
$P = 0.035$ (Figure~\ref{fig:pillar1}c,d). All nine estimates favour PathOPEN
($|r|$ from 0.09 to 0.40). In the six judge-only comparisons with Quilt-VQA,
scores were also higher for PathOPEN and every interval excluded zero
($P \le 4 \times 10^{-4}$; Figure~\ref{fig:pillar1}f).

\paragraph{Variation across pathologist pools.}
Pathologists' mean scores varied less for PathOPEN than for filtered PathVQA
(Figure~\ref{fig:pillar1}g, Table~\ref{tab:per-evaluator}). The spread was about
five times smaller on the open-ended criteria and 24 times smaller on the
close-ended criterion. Kruskal--Wallis tests found differences across the
PathOPEN pools on the two Benchmark~1 criteria ($P < 0.001$, $n = 686$)
and gave $P = 0.43$ for Benchmark~3. Maximum scores made up 84--100\% of each
pathologist's PathOPEN ratings, leaving little room for their means to differ.
The separate item pools also affect these comparisons (Section~\ref{subsubsec:stats}).

\paragraph{Judge--pathologist agreement.}
On PathOPEN Benchmark~3, Qwen matched pathologist scores on 95.2\% of items.
Quadratic-weighted Cohen's $\kappa$ was $-0.022$ and Gwet's AC1 was 0.984
(Figure~\ref{fig:pillar1}e). Roughly 97\% of ratings fell in one category.
With that concentration, Cohen's $\kappa$ estimates a high expected agreement
by chance even when the observed agreement is high.

\PathOPENFigure{\includegraphics[width=\linewidth,height=\pathopenfigureheight,keepaspectratio]{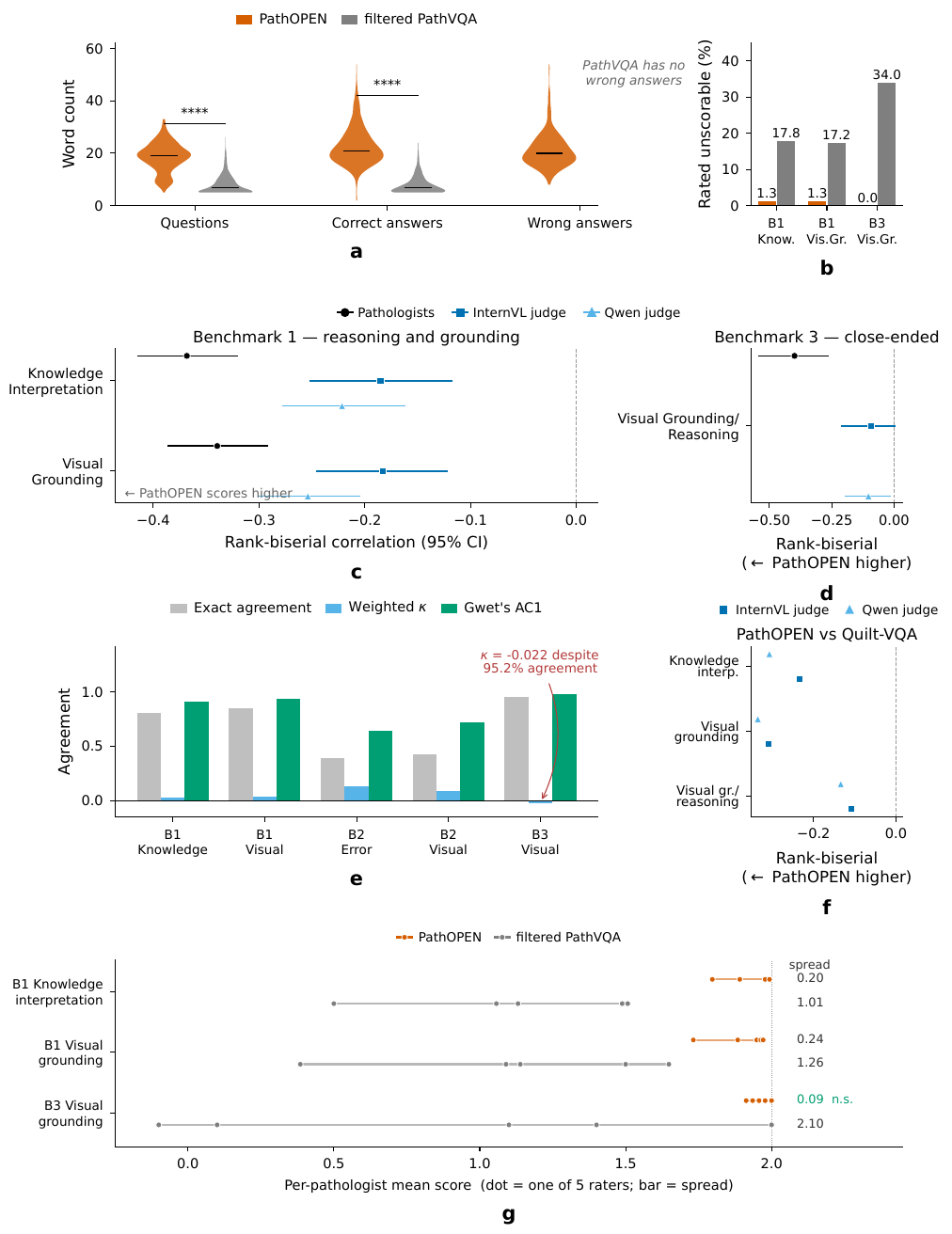}}{\textbf{Ratings of pathology question--answer pairs.}
\textbf{a}, Word-count distributions for questions, correct answers and wrong answers;
violins show the distributions and lines mark medians. Four asterisks denote
$P < 10^{-4}$ in two-sided Mann--Whitney $U$ tests. Filtered PathVQA provides no
matched open-ended wrong answers. \textbf{b}, Percentage of pathologist ratings scored $-1$ (unable to evaluate) per
criterion. The PathOPEN denominators are 231 for each Benchmark~1 criterion and 230
for Benchmark~3.
\textbf{c}, \textbf{d}, Rank-biserial effects against filtered PathVQA for
Benchmarks~1 and 3, respectively. In \textbf{c}, the first native PathOPEN open-ended
slot contributes 231 human ratings per criterion, against 624 reasoning and 622 grounding
ratings for PathVQA. Each judge scored 229 versus 612 items. The historical PathVQA totals
come from the Benchmark~1 score columns. These include 12
reasoning and 10 grounding scores recorded on close-ended rows
(Section~\ref{subsubsec4_5_2}). In \textbf{d}, the human comparison uses
230 versus 50 items and each judge comparison uses 229 versus 50.
\textbf{e}, Exact agreement, quadratic-weighted Cohen's $\kappa$ and Gwet's AC1 between
the Qwen judge and the pathologist who rated each item. \textbf{f}, Effects against
Quilt-VQA, with 229 versus 724 open-ended items and 229 versus 257 yes/no items per judge. \textbf{g}, Mean scores across disjoint pathologist
pools: dots mark means, bars their ranges and adjacent values the spreads;
\emph{n.s.} denotes Kruskal--Wallis $P \ge 0.05$. Effects have bootstrap 95\% confidence
intervals (2{,}000 resamples), with negative values indicating higher PathOPEN scores;
the dashed line marks zero. Encoded $-1$ values remain in these summaries.
Section~\ref{sec:pillar1} gives the denominators for pathologists and each judge.}{fig:pillar1}

\subsection{Graded difficulty structure of the wrong answers}
\label{sec:pillar2}

Benchmark~2 groups PathOPEN's wrong answers by how close they come to a correct
answer and how they use the image. It covers 347 open-ended wrong-answer pairs
with an available image and the distractors from the multiple-choice questions
(Section~\ref{subsec:qa-records}). Comparator MCQ distractors are assessed under
Benchmark~5 (Section~\ref{sec:pillar3}).

\paragraph{Tier definitions and distributions.}
Scores of 2, 1 and 0 map to near-miss, moderate-miss and far-miss tiers, respectively.
A score of $-1$ carries no tier. The historical sheets contain 349 native wrong answers,
including two linked to image slots without files in the released package. Of the 698
possible ratings for these answers, two are blank and 12 are $-1$, leaving 684 tier
ratings. The sheets also contain two scores for a slot with no wrong-answer text.
Those scores fall outside the answer-linked count. For MCQ, 920 distractors have
recorded scores on both criteria. Nine of those 1{,}840 scores are $-1$, leaving
1{,}831 tier ratings.

Pathologists gave open-ended wrong answers 116 near-miss, 300 moderate-miss and
268 far-miss ratings (17.0\%, 43.9\% and 39.2\%). MCQ distractors received 257,
465 and 1{,}109 ratings in those tiers (14.0\%, 25.4\% and 60.6\%).
The near and moderate tiers together accounted for 60.8\% of open-ended ratings
and 39.4\% of distractor ratings. We matched each judge's scores with the
pathologist's scores by image and answer slot. Figure~\ref{fig:pillar2}a
shows the pathologist distributions for the 668 native and 1{,}773 MCQ score pairs
with a tier from both the pathologist and Qwen. The corresponding InternVL comparison
contains 680 native and 1{,}815 MCQ score pairs.

\paragraph{Judge agreement with pathologist tiers.}
Agreement with pathologists was low. Across the 24 combinations of judge,
criterion and answer slot, every quadratic-weighted $\kappa$ was below 0.26 and
one was negative. Mean $\kappa$ was 0.114 for InternVL and 0.142 for Qwen
(Table~\ref{tab:judge-calibration} summarises agreement by answer type).
The difficulty analyses use the pathologists' labels.

Both judges assigned the near-miss tier more often than pathologists. InternVL
assigned it to 43.2\% of ratings, compared with 14.6\% for pathologists. Its
near-miss rates for items pathologists labelled near-miss and moderate-miss
were similar. Agreement was lower for native open-ended wrong answers than
for MCQ distractors for both judges (Table~\ref{tab:judge-calibration}).

The confusion matrices use only the Error Proximity and Deductive Plausibility
criterion, pooling two open-ended answer slots and four MCQ distractor slots.
They contain 1{,}248 InternVL and 1{,}221 Qwen score pairs. The matrices show 57.9\% of InternVL assignments and 41.0\% of Qwen assignments
below the diagonal, where a judge assigns a harder tier than the pathologist; exact
tier matches account for 32.7\% and 41.2\%, respectively (Figure~\ref{fig:pillar2}b).
Calibration of the judge scores remains to be evaluated.

\paragraph{Behavioural checks of tier difficulty.}
Five of six models selected near-miss distractors more often than far-miss
distractors after dividing by the number offered in each tier
(Figure~\ref{fig:pillar2}c). Models often chose moderate-miss distractors most
frequently, and only Patho-R1 showed an ordered trend across all three tiers.
With three tiers, the smallest possible $P$ for the Kendall $\tau$ test is
0.333. This comparison describes how models behave when they see the distractors
as answer choices.

We also checked whether item-level accuracy was lower when an item's hardest
distractor was a near-miss. That direction occurs for 2 of 6 models. Pathologist scores on the first Benchmark~2 criterion gave difficulty tiers for
343 historical open-ended image--question pairs. All six models had answers for 341 of these pairs. The
two unmatched pairs were the two question slots for case~79. Across the 341
matched pairs, mean token F1 was 22.3 for near
misses, 21.4 for moderate misses and 22.0 for far misses. Three of six models
followed the expected direction. These two checks show little consistency across
models. Item accuracy also depends on the question, and in open-ended answering
the model sees the question and image without the reference wrong answer.

\subsection{Standalone assessment of the MCQ options}
\label{sec:pillar3}

Benchmark~5 assesses the MCQ options as standalone clinical statements. Benchmark~5a
rates correct options for Standalone Completeness and Visual Grounding; Benchmark~5b
rates distractors for Distractor Standalone Plausibility and Distractor Visual Grounding
Error. InternVL and Qwen each
made 6{,}971 calls across PathOPEN, PathMMU and PatchVQA. This benchmark uses judges
only; their agreement measures consistency between models, with no pathologist baseline.

PathOPEN's authoring protocol requested complete clinical statements. Its 1{,}145
evaluated options average 9.67 words, against 4.81 for PathMMU and 4.98 for PatchVQA;
medians are 11, 4 and 4. PathOPEN option text is therefore roughly twice as long in this
evaluation pool.

\paragraph{Correct-option completeness and grounding.}
Both judges give PathOPEN the highest Standalone Completeness mean.
InternVL scores PathOPEN 1.891, PathMMU 1.686 and PatchVQA 1.480; Qwen scores them
1.712, 0.863 and 0.728, respectively (Figure~\ref{fig:pillar3}a). InternVL also assigned the maximum completeness score more often to PathOPEN
(215/229) than to PathMMU (741/905) or PatchVQA (207/294)
(Figure~\ref{fig:pillar3}c). For visual grounding, InternVL gave similar scores to all three datasets
(1.852 for PathOPEN, 1.885 for PathMMU and 1.810 for PatchVQA).
Neither judge consistently favoured PathOPEN on this criterion.

\paragraph{Distractor ratings differ between judges.}
The judges rank the datasets oppositely on both Benchmark~5b criteria under the same
rubric and on the same items (Figure~\ref{fig:pillar3}b). InternVL places PathOPEN first,
with Distractor Standalone Plausibility 1.19 and Distractor Visual Grounding Error 0.85. Qwen places it last, with 1.42
and 0.60, behind PatchVQA's 1.55 and 0.81. Across the four Benchmark~5 dimensions,
PathOPEN ranks highest on three under InternVL and one under Qwen. The distractor
comparison is therefore judge-dependent.

Generation truncation accounts for $-1$ on 2.3--3.5\% of Qwen's distractor-criterion
scores, compared with 0.1--0.4\% on correct-option criteria and none for InternVL.
The reported effect on a mean is at most 0.09, insufficient to explain the opposite
dataset orderings. Mean judge--judge quadratic-weighted $\kappa$ is 0.371 for PathOPEN,
0.265 for PatchVQA and 0.224 for PathMMU (Figure~\ref{fig:pillar3}e,f). PathOPEN has the
highest Benchmark~5a agreement, at 0.547 and 0.307; agreement is weakest overall on
PathMMU, and no cell reaches substantial agreement. We report the judges separately because they agree on completeness but
rank the distractors differently.

\paragraph{Answering with all options.}
Figure~\ref{fig:pillar3}d shows the separate task in which the six answering models
see the image and all options. Section~\ref{sec:pillar4} reports their accuracies
and the chance level for each dataset.

\subsection{Benchmarking utility across question formats}
\label{sec:pillar4}

We tested six answering models on multiple-choice, close-ended and open-ended
questions without task-specific examples. Tables~\ref{tab:dataset-comparison}
and \ref{tab:format-ranks} report confidence intervals, pairwise tests and rank
correlations. The bootstrap uses individual model--item pairs
(Section~\ref{subsubsec:stats}).

\paragraph{Multiple-choice performance.}
With the image and all options present, mean raw accuracy is 42.4\% for PathOPEN,
50.1\% for PathMMU and 41.4\% for PatchVQA
(Figures~\ref{fig:pillar3}d and \ref{fig:pillar4}a).
Option counts range from 2 to 16 across datasets. PathOPEN has five options and a
20.0\% chance level, compared with PathMMU's 24.3\%; accuracy above each dataset's
chance level accounts for this difference in guessing baselines.

PathOPEN averaged 22.36 percentage points above chance, between PathMMU at 25.75
and PatchVQA at 20.46. The ordering by mean accuracy was the same before and after adjustment.
Table~\ref{tab:dataset-comparison} gives the confidence intervals for these means
and the separate rank comparisons of adjusted scores across model--item pairs. PathOPEN
has the narrowest performance range across the six models, 28.4 points, compared with
39.8 for PathMMU.

\paragraph{Close-ended and open-ended performance.}
On yes/no questions, PathOPEN averaged 62.66\% accuracy, between Quilt-VQA at
66.21\% and filtered PathVQA at 55.00\%. All three exceed the 50\% chance level.
Pairwise differences remain significant after Holm correction, though PathOPEN's
effects against filtered PathVQA (0.077) and Quilt-VQA ($-0.035$) are small
(Figure~\ref{fig:pillar4}b, Table~\ref{tab:dataset-comparison}).

Table~\ref{tab:dataset-comparison} reports open-ended token F1 multiplied by 100 on
the full reference set. A separate analysis restricts reference length to 16--35 words to
reduce length differences: the full evaluation references average 22.8 words for
PathOPEN and 7.8 for filtered PathVQA. With references restricted to this length range, all six models score higher
on PathOPEN than on filtered PathVQA and lower than on Quilt-VQA. The pooled
ordering is consistent across the metrics reported in this comparison.

Figure~\ref{fig:pillar4}c shows BERTScore F1 rescaled against an empirical baseline,
where zero means similarity no greater than unrelated text. Filtered PathVQA falls
below zero for two of six models (Quilt-LLaVA $-5.8$, Patho-R1 $-1.0$) and below 1.5
for four. Short reference answers give fewer tokens with which the model's answer can
overlap. Interpret scores in light of both reference length and answer quality.

\paragraph{Model rankings across formats.}
Patho-R1 ranks first on MCQ and fifth on open-ended questions; LLaVA-Med ranks fifth
on MCQ and second on both other formats. Only InternVL retains its position across
all three formats (Figure~\ref{fig:pillar4}d). The three observed format-pair Spearman
correlations are non-significant among the six models (Table~\ref{tab:format-ranks}).
For MCQ versus open-ended performance, $\rho = 0.20$ after averaging over all three
datasets and $\rho = 0.54$ on PathOPEN alone, with $P > 0.25$ for both.
These observed rank changes support reporting formats separately; the six-model
tests leave uncertainty about the relationship between rankings, and estimates
depend on which datasets enter the average.

\paragraph{Magnification and organ-system strata.}
Across 18 model-by-format tests, two magnification associations reach uncorrected
$P < 0.05$: Patho-R1 on MCQ ($P = 0.003$) and Quilt-LLaVA on open-ended questions
($P = 0.009$; Figure~\ref{fig:pillar4}e). The two results are specific to those model and format combinations. Some
magnification groups have few items. These tests use microscope magnification
recorded in the metadata, separately from image dimensions in pixels.

Figure~\ref{fig:pillar4}f--k gives performance by magnification and organ system for
each format, with item counts printed per bin and bins below $n = 30$ marked.
The comparator records used here lack metadata for the same analysis. PathOPEN has
one or two open-ended questions per image, against one MCQ and one close-ended
question. Its open-ended results have 1 of 6 magnification bins and 5 of 13 organ
systems below $n = 30$, compared with 3 of 6 and 10 of 11 for MCQ. The larger
open-ended counts provide more item-level observations, with questions and images
within a case still dependent.

\subsection{Augmented-image quality and class-balanced retention}
\label{sec:pillar5}

Benchmark~4 asks whether augmented images still support the original answers.
We also measured how selecting more augmentations from rare classes changes the
representation of each metadata field.

\paragraph{Pathologist and judge ratings of augmented images.}
Pathologists and both judges rated the same 174 images in the historical Benchmark~4
evaluation set (Section~\ref{subsubsec4_5_3}).
Five pathologists assign the maximum Benchmark~4 score to 88.3\% of ratings for
Clinical Relevance and 84.2\% for Visual Grounding. The two judges' maximum-score
rates range from 89.1\% to 91.1\% (Figure~\ref{fig:pillar5}a). The pathologist
percentages include unscorable ratings in their denominators. Across pathologists,
both criteria differ at Kruskal--Wallis $P < 0.001$, with mean-score spreads of
approximately 0.5, compared with 0.20--0.24 on original PathOPEN items.
Pathologist~5 marks 12.2\% and 9.5\% of augmented images unscorable on the two criteria.
The other four mark none (Table~\ref{tab:benchmark4-rating}).

Exact agreement between judges and pathologists ranges from 75.0\% to 83.8\%,
and AC1 ranges from 0.910 to 0.947. With ratings concentrated at the maximum,
these measures help interpret $\kappa$: seven of its eight confidence
intervals include zero. InternVL's
visual-grounding interval for the first question slot has a lower bound of 0.001.
Table~\ref{tab:benchmark4-rating} gives exact agreement, $\kappa$ and AC1 for both
judges and both question slots.

Benchmark~4 was applied only to augmented images. Figure~\ref{fig:pillar5}b compares
their Visual Grounding scores with Benchmark~1 Visual Grounding on originals, using
unpaired pools and differently worded criteria. The signed effects differ across
rater sources: pathologists $-0.093$, InternVL $+0.064$ and Qwen $-0.003$.
The estimates compare groups rated under two rubrics. Estimating change from
each original would require rating both images under the same rubric.

\paragraph{Class balance in the selected subsets.}
Applying 30 augmentations to every image preserves the original class proportions.
The four manifests change those proportions by retaining up to ten augmentations
per source image, with more files from rare classes
(Section~\ref{subsubsec:balancing}). Section~\ref{subsec:augmentation-records}
lists the four manifest sizes at the default cap. The larger categorization
selection of 1{,}725 images comes from the analysis at a cap of 30.

The largest-to-smallest class ratio falls from 77.0$\times$ to 8.2$\times$ for
categorization, 25.0$\times$ to 2.8$\times$ for organ, 10.4$\times$ to 1.1$\times$ for
magnification and 9.0$\times$ to 1.5$\times$ for regional anatomy
(Figure~\ref{fig:pillar5}c,d). Gini falls and normalised entropy rises for all four
fields (Figure~\ref{fig:pillar5}e,f). Under this rule, each image in the largest class
contributes one augmentation, so the largest categorization class still contributes
77 files and its residual ratio is 8.2$\times$ at the default cap.

Each manifest balances one field, and their selections can overlap. Regional anatomy
nests within organ, so balancing one field also changes others; no single variant
balances all four simultaneously.

\paragraph{Effect of the retention cap.}
Magnification and regional anatomy reach their lowest Gini at a cap of ten
augmentations per source image and remain unchanged at higher caps (Figure~\ref{fig:pillar5}g,h). Organ and categorization
improve further as the selected sets grow. Even at 30$\times$, the organ-balanced set
contains 1{,}036 images, averaging 4.5 per source, because frequent classes receive
small multipliers under the selection rule. Increasing the cap changes class
balance and the number of derived views per source image. These comparisons
describe those two properties. Training and overfitting would need separate tests.

\sbox{\pathopentablebox}{%
\begin{tabular}{l l S[table-format=2.1] S[table-format=2.1] S[table-format=2.1] S[table-format=2.1] S[table-format=2.1] S[table-format=2.1] S[table-format=3.0]}
\toprule
Dataset & Criterion & \multicolumn{5}{c}{\% items scored $-1$, per pathologist} & {Range} & {Pooled} \\
\cmidrule(lr){3-7}
 & & {P1} & {P2} & {P3} & {P4} & {P5} & & {$n_{-1}$} \\
\midrule
PathOPEN & B1 Knowledge interpretation & 0.0 & 0.0 & 0.0 & 0.0 & 3.6 & 3.6 & 5 \\
PathOPEN & B1 Visual grounding & 0.0 & 0.0 & 0.0 & 0.0 & 3.6 & 3.6 & 5 \\
PathOPEN & B3 Visual grounding & 0.0 & 0.0 & 0.0 & 0.0 & 0.0 & 0.0 & 0 \\
\midrule
filtered PathVQA & B1 Knowledge interpretation & 18.7 & 40.2 & 13.1 & 3.3 & 14.2 & 36.9 & 111 \\
filtered PathVQA & B1 Visual grounding & 18.9 & 40.2 & 12.3 & 2.5 & 12.7 & 37.7 & 107 \\
filtered PathVQA & B3 Visual grounding & 0.0 & 60.0 & 70.0 & 10.0 & 30.0 & 70.0 & 17 \\
\bottomrule
\end{tabular}
}

\begin{tableorg}[t]
\centering\small
\caption{\textbf{Unscorable-item rates across pathologist pools.}
Entries give the percentage of items scored $-1$ (unable to evaluate) by each pathologist
(P1--P5). Range is the largest minus the smallest percentage, in percentage points;
$n_{-1}$ is the pooled count of unscorable ratings. B1 and B3 denote Benchmarks~1 and 3;
their full criterion names are given in Supplementary Tables~\ref{tab:benchmark1} and
\ref{tab:benchmark3}. The disjoint item pools combine rater and item-composition
differences; $-1$ does not identify a single cause (Section~\ref{sec:pillar1}).}
\label{tab:defect-mechanism}
\resizebox{\linewidth}{!}{\usebox{\pathopentablebox}}
\end{tableorg}

\sbox{\pathopentablebox}{%
\begin{tabular}{l l S[table-format=4.0] S[table-format=-1.3] S[table-format=-1.3] S[table-format=-1.3] S[table-format=-1.3] S[table-format=-1.3] S[table-format=1.3] c}
\toprule
Dataset & Criterion & {$n$} & \multicolumn{5}{c}{Pathologist mean score} & {Spread} & {$P$} \\
\cmidrule(lr){4-8}
 & & & {P1} & {P2} & {P3} & {P4} & {P5} & & \\
\midrule
PathOPEN & B1 Knowledge interpretation & 686 & 1.891 & 1.978 & 1.978 & 1.993 & 1.797 & 0.196 & $<$0.001 \\
PathOPEN & B1 Visual grounding & 686 & 1.884 & 1.949 & 1.963 & 1.971 & 1.732 & 0.239 & $<$0.001 \\
PathOPEN & B3 Visual grounding & 230 & 2.000 & 1.913 & 1.978 & 1.935 & 1.957 & 0.087 & \textbf{n.s.} \\
\midrule
filtered PathVQA & B1 Knowledge interpretation & 624 & 1.057 & 0.500 & 1.131 & 1.488 & 1.507 & 1.007 & $<$0.001 \\
filtered PathVQA & B1 Visual grounding & 622 & 1.090 & 0.385 & 1.139 & 1.648 & 1.500 & 1.262 & $<$0.001 \\
filtered PathVQA & B3 Visual grounding & 50 & 2.000 & 0.100 & -0.100 & 1.400 & 1.100 & 2.100 & 0.004 \\
\bottomrule
\end{tabular}
}

\begin{tableorg}[t]
\centering\small
\caption{\textbf{Mean scores across pathologist pools.}
P1--P5 identify the five pathologists, each rating a disjoint item pool; $n$ is the pooled
rating count per criterion. Means and Kruskal--Wallis tests retain encoded $-1$ values
(Section~\ref{subsubsec:stats}). Spread is the range of the five means; the test compares
their score distributions, with \emph{n.s.} denoting $P \ge 0.05$. B1 and B3 denote
Benchmarks~1 and 3, whose full criterion names appear in Supplementary
Tables~\ref{tab:benchmark1} and \ref{tab:benchmark3}. These separate pools do not support
per-item inter-rater coefficients (Section~\ref{sec:pillar1}).}
\label{tab:per-evaluator}
\resizebox{\linewidth}{!}{\usebox{\pathopentablebox}}
\end{tableorg}

\begin{table}[t]
\centering\small
\caption{\textbf{Judge and pathologist assignments of wrong-answer difficulty tiers.}
The upper and middle blocks use Error Proximity and Deductive Plausibility,
with 1{,}248 InternVL and 1{,}221 Qwen score pairs. Both near-miss ratios use
the pathologists' frequency in the InternVL-matched pool as their denominator.
The middle block gives each judge's distribution conditional on the pathologist tier.
The lower block summarises quadratic-weighted $\kappa$ across both criteria
and six answer slots per judge. By authoring route, OE-native denotes native open-ended wrong answers, and MCQ-derived
denotes multiple-choice question distractors. Near, Moderate and Far abbreviate
near-miss, moderate-miss and far-miss. Scores of $-1$ carry no tier.
Section~\ref{sec:pillar2} describes the evaluation pools and agreement.}
\label{tab:judge-calibration}
\resizebox{\linewidth}{!}{%
\begin{tabular}{l l S[table-format=-2.3] S[table-format=-2.3] S[table-format=-2.3] S[table-format=1.2]}
\toprule
 & & \multicolumn{3}{c}{Tier assigned (\%)} & {Near-miss} \\
\cmidrule(lr){3-5}
Rater & Split & {Near} & {Moderate} & {Far} & {inflation} \\
\midrule
Pathologists & InternVL-matched & 14.6 & 34.9 & 50.5 & {--} \\
InternVL judge & InternVL-matched & 43.2 & 42.6 & 14.2 & 2.96 \\
Qwen judge & Qwen-matched & 35.5 & 24.1 & 40.4 & 2.44 \\
\midrule
\multicolumn{6}{l}{\emph{Per-tier recall: what the judge said, given what the pathologist said}} \\
\midrule
InternVL judge & path.\ said near-miss & 52.7 & 39.6 & 7.7 & {--} \\
InternVL judge & path.\ said moderate-miss & 51.1 & 41.5 & 7.3 & {--} \\
InternVL judge & path.\ said far-miss & 34.9 & 44.3 & 20.8 & {--} \\
Qwen judge & path.\ said near-miss & 44.9 & 25.3 & 29.8 & {--} \\
Qwen judge & path.\ said moderate-miss & 46.7 & 24.6 & 28.7 & {--} \\
Qwen judge & path.\ said far-miss & 25.4 & 23.4 & 51.2 & {--} \\
\midrule
\multicolumn{6}{l}{\emph{Weighted $\kappa$ against the pathologists, by source of the distractor}} \\
\cmidrule(lr){3-5}
 & & {Mean} & {Min} & {Max} & {Cells} \\
\midrule
InternVL judge & OE-native (free text) & 0.054 & -0.021 & 0.100 & {4} \\
InternVL judge & MCQ-derived (option) & 0.144 & 0.111 & 0.181 & {8} \\
Qwen judge & OE-native (free text) & 0.104 & 0.065 & 0.145 & {4} \\
Qwen judge & MCQ-derived (option) & 0.161 & 0.079 & 0.253 & {8} \\
\bottomrule
\end{tabular}%
}
\begin{flushleft}\footnotesize Mean, minimum and maximum are taken across that judge's criteria and distractor slots.\end{flushleft}
\end{table}

\sbox{\pathopentablebox}{%
\begin{tabular}{l l S[table-format=4.0] S[table-format=-2.3] c}
\toprule
Format & Dataset & {Items} & {Mean} & {95\% CI} \\
\midrule
MCQ (accuracy) & PathMMU & 5430 & 50.06 & [48.71, 51.36] \\
 & PathOPEN & 1374 & 42.36 & [39.88, 44.98] \\
 & PatchVQA & 1764 & 41.38 & [39.06, 43.88] \\
\addlinespace
MCQ (above chance) & PathMMU & 5430 & 25.75 & [24.42, 27.04] \\
 & PathOPEN & 1374 & 22.36 & [19.88, 24.98] \\
 & PatchVQA & 1764 & 20.46 & [18.14, 22.91] \\
\addlinespace
Close-ended & Quilt-VQA & 1542 & 66.21 & [63.81, 68.61] \\
 & PathOPEN & 1374 & 62.66 & [60.04, 65.14] \\
 & filtered PathVQA & 300 & 55.00 & [49.33, 60.67] \\
\addlinespace
Open-ended & Quilt-VQA & 4344 & 26.56 & [26.11, 27.01] \\
 & PathOPEN & 2718 & 21.90 & [21.54, 22.26] \\
 & filtered PathVQA & 3672 & 8.66 & [8.40, 8.93] \\
\addlinespace
\midrule
\multicolumn{5}{l}{\emph{Pairwise comparisons (Mann--Whitney $U$, Holm-corrected within format)}} \\
\midrule
Format & Comparison & {} & {Rank-biserial} & {$P_{\mathrm{Holm}}$} \\
\midrule
MCQ (above chance) & PatchVQA vs PathMMU & & 0.081 & $<$0.001 \\
MCQ (above chance) & PatchVQA vs PathOPEN & & -0.053 & 0.010 \\
MCQ (above chance) & PathMMU vs PathOPEN & & -0.317 & $<$0.001 \\
Close-ended & PathOPEN vs filtered PathVQA & & 0.077 & 0.027 \\
Close-ended & PathOPEN vs Quilt-VQA & & -0.035 & 0.046 \\
Close-ended & filtered PathVQA vs Quilt-VQA & & -0.112 & $<$0.001 \\
Open-ended & PathOPEN vs filtered PathVQA & & 0.740 & $<$0.001 \\
Open-ended & PathOPEN vs Quilt-VQA & & -0.161 & $<$0.001 \\
Open-ended & filtered PathVQA vs Quilt-VQA & & -0.768 & $<$0.001 \\
\bottomrule
\end{tabular}
}

\begin{tableorg}[t]
\centering\small
\caption{\textbf{Pooled zero-shot performance across datasets.}
Means pool six answering models, treating each (model, item) pair as one observation
in the Items column. Percentile bootstrap 95\% confidence intervals (CIs) use 2{,}000
resamples of those observations. Multiple-choice question (MCQ) and close-ended
accuracies are percentages; above-chance accuracy is a percentage-point difference
from each dataset's chance level. Open-ended scores are token F1 multiplied by 100,
using the full reference set. Rank-biserial correlations are positive when the
first named dataset tends to rank higher. For above-chance MCQ scores the tests
compare the ranks of individual model--item scores after subtracting each
item's guessing baseline. The order of these ranks can differ from the order
of the dataset means. Mann--Whitney $U$ tests are Holm-corrected within
format. The bootstrap treats repeated items and within-case observations as independent,
so intervals may understate uncertainty from those dependencies
(Sections~\ref{subsubsec:stats} and \ref{sec:pillar4}).}
\label{tab:dataset-comparison}
\resizebox{\linewidth}{!}{\usebox{\pathopentablebox}}
\end{tableorg}

\begin{table}[t]
\centering\small
\caption{\textbf{Model rank correlations across question formats.}
Spearman correlations compare the six models' mean performance on PathOPEN for each
pair of formats. MCQ denotes multiple-choice question. None of the three observed
correlations reaches $P < 0.05$; six-model estimates leave uncertainty about the
relationship between rankings. Section~\ref{sec:pillar4} also reports the MCQ--open-ended
correlation after averaging across datasets.}
\label{tab:format-ranks}
\begin{tabular}{l S[table-format=1.0] S[table-format=-1.3] c c}
\toprule
Format pair & {Models} & {Spearman $\rho$} & {$P$} & {Significant} \\
\midrule
MCQ vs Open-ended & 6 & 0.543 & 0.266 & no \\
MCQ vs Close-ended & 6 & 0.257 & 0.623 & no \\
Close-ended vs Open-ended & 6 & 0.257 & 0.623 & no \\
\bottomrule
\end{tabular}
\end{table}

\begin{table}[t]
\centering\small
\caption{\textbf{Pathologist ratings and judge--pathologist agreement on augmented images.}
The upper block gives each pathologist's mean score, maximum-score percentage and
unscorable percentage under Benchmark~4; P1--P5 identify the five pathologists. Spread
is the range of their means, and $P$ is the Kruskal--Wallis test across their disjoint
pools. These summaries retain encoded $-1$ values. The lower block gives exact
agreement, quadratic-weighted Cohen's $\kappa$ with 95\% confidence intervals (CIs),
and Gwet's AC1 on shared judge--pathologist items. The three statistics describe
agreement under the concentrated score distributions (Section~\ref{sec:pillar5}).}
\label{tab:benchmark4-rating}
\begin{tabular}{l l r l r}
\toprule
Criterion & Rater & {Mean} & {\% max} & {\% unscorable} \\
\midrule
Clinical Relevance & P1 & 1.92 & 93.9 & 0.0 \\
 & P2 & 1.91 & 92.5 & 0.0 \\
 & P3 & 1.96 & 95.5 & 0.0 \\
 & P4 & 1.87 & 86.6 & 0.0 \\
 & P5 & 1.43 & 74.3 & 12.2 \\
 & \emph{spread / $P$} & 0.52 & \multicolumn{2}{c}{$<$0.001} \\
\addlinespace
Visual Grounding & P1 & 1.96 & 97.0 & 0.0 \\
 & P2 & 1.85 & 86.6 & 0.0 \\
 & P3 & 1.94 & 94.0 & 0.0 \\
 & P4 & 1.64 & 70.1 & 0.0 \\
 & P5 & 1.46 & 74.3 & 9.5 \\
 & \emph{spread / $P$} & 0.49 & \multicolumn{2}{c}{$<$0.001} \\
\addlinespace
\midrule
\multicolumn{5}{l}{\emph{Judge versus pathologist on the same augmented images}} \\
\midrule
Criterion & Judge, slot & {Exact \%} & {$\kappa$ [95\% CI]} & {AC1} \\
\midrule
Clinical Relevance & InternVL, 1 & 83.8 & 0.12 [-0.06, 0.33] & 0.942 \\
 & InternVL, 2 & 81.0 & 0.09 [-0.09, 0.26] & 0.947 \\
 & Qwen, 1 & 82.1 & 0.09 [-0.09, 0.32] & 0.923 \\
 & Qwen, 2 & 83.3 & 0.18 [-0.06, 0.43] & 0.939 \\
\addlinespace
Visual Grounding & InternVL, 1 & 80.9 & 0.24 [0.001, 0.472] & 0.941 \\
 & InternVL, 2 & 75.0 & 0.01 [-0.11, 0.15] & 0.919 \\
 & Qwen, 1 & 79.8 & 0.03 [-0.11, 0.21] & 0.910 \\
 & Qwen, 2 & 79.8 & 0.14 [-0.04, 0.35] & 0.918 \\
\bottomrule
\end{tabular}
\begin{flushleft}\footnotesize Slots 1 and 2 are the two augmented open-ended question slots. The agreement estimates use $n=173$ and $n=168$ paired scores, respectively.\end{flushleft}
\end{table}

\PathOPENFigure{\includegraphics[width=\linewidth,height=\pathopenfigureheight,keepaspectratio]{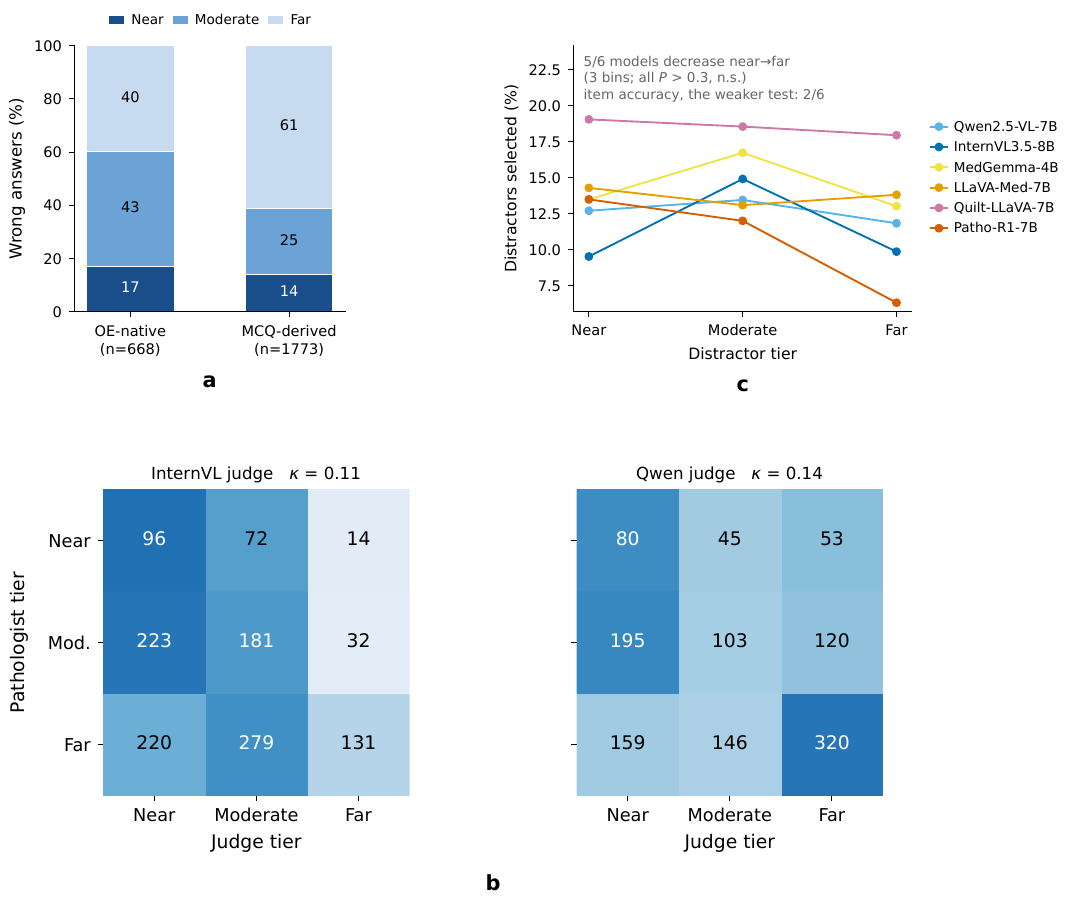}}{\textbf{Wrong-answer difficulty tiers, judge agreement and distractor attraction.}
\textbf{a}, Pathologist-assigned tiers for native open-ended answers and multiple-choice
question (MCQ) distractors. Both Benchmark~2 criteria are pooled. The denominators
are 668 native and 1{,}773 MCQ score pairs with a tier from both the pathologist and
Qwen.
\textbf{b}, Confusion matrices for Error Proximity and Deductive Plausibility alone,
with 1{,}248 InternVL and 1{,}221 Qwen score pairs across both answer types. Rows give
the pathologist tier and columns the judge tier. Below the diagonal, the judge
assigned a harder tier.
\textbf{c}, Distractor attraction, measured as the percentage of offered distractors
selected within each tier. Five of six models select near-miss distractors more often
than far-miss distractors. Item-level accuracy, annotated in the panel, follows the
expected direction for only 2 of 6 models. Benchmark~2 scores of 2, 1 and 0 define
near-miss, moderate-miss and far-miss tiers; $-1$ has no tier.
Section~\ref{sec:pillar2} and Table~\ref{tab:judge-calibration} give counts, agreement
coefficients, full-pool results and the limits of the three-bin significance test.}{fig:pillar2}

\PathOPENFigure{\includegraphics[width=\linewidth,height=\pathopenfigureheight,keepaspectratio]{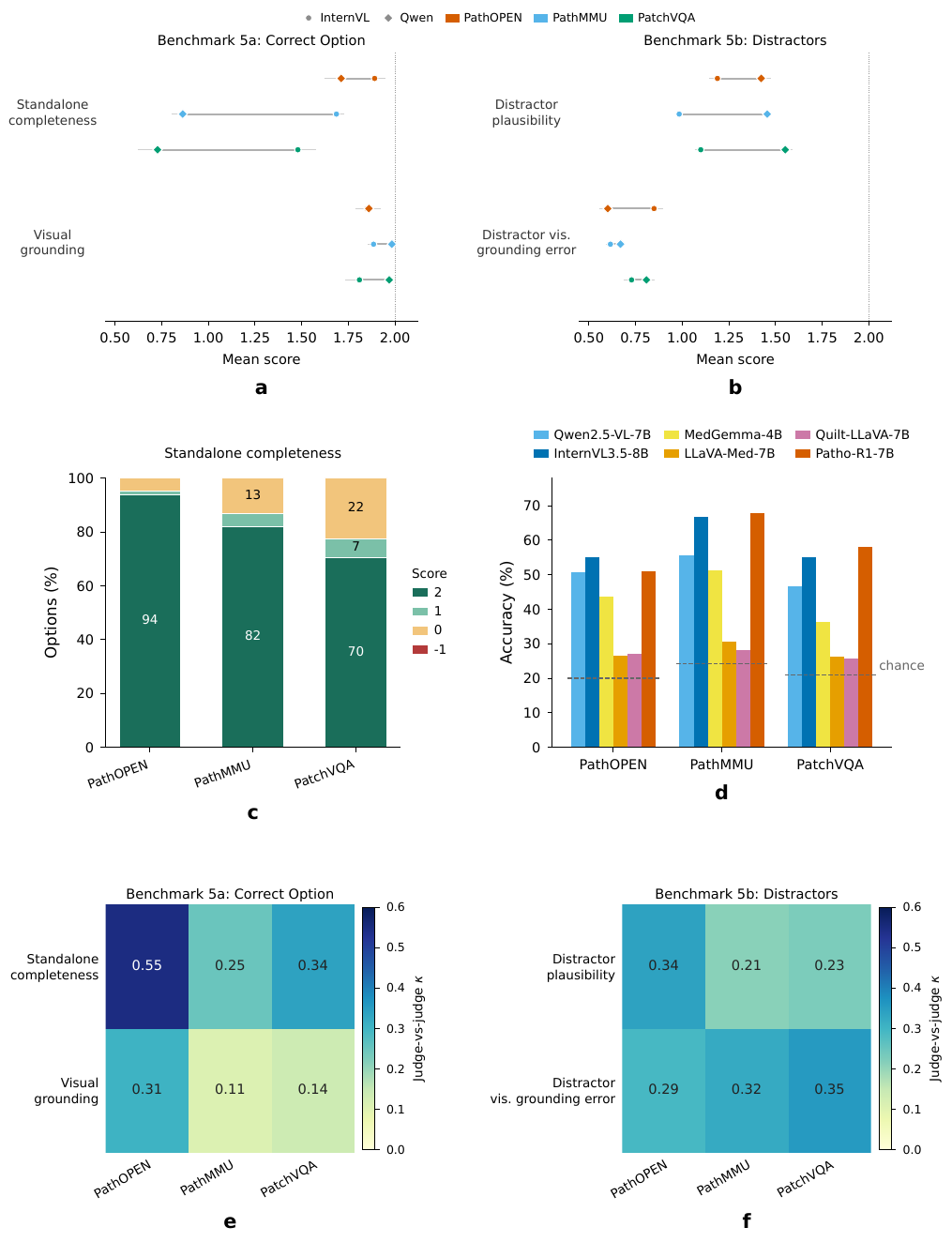}}{\textbf{Judge assessment of MCQ options and full-context answering.}
\textbf{a}, \textbf{b}, Mean Benchmark~5 scores by criterion and dataset for correct
options (5a) and distractors (5b). Circles denote InternVL and diamonds Qwen; connectors
show their score differences, colours identify datasets and whiskers show bootstrap
95\% confidence intervals. The judges reverse their dataset ordering on both
distractor criteria, so their results are shown separately. \textbf{c}, Standalone
Completeness score distributions for InternVL. \textbf{d}, Zero-shot multiple-choice question (MCQ)
accuracy with the image and all options present; dashed lines indicate dataset-specific
chance levels for option counts from 2 to 16. \textbf{e}, \textbf{f}, Judge--judge
quadratic-weighted $\kappa$ for 5a and 5b, with separate scales. Benchmark~5 has no
pathologist baseline: these coefficients measure agreement between the two judges.
Short criterion labels in the panels refer to the full rubrics in Supplementary
Tables~\ref{tab:benchmark5a} and \ref{tab:benchmark5b}.
Sections~\ref{sec:pillar3} and \ref{sec:pillar4} report the option assessments and
full-context answering results, respectively.}{fig:pillar3}

\PathOPENFigure{\includegraphics[width=\linewidth,height=\pathopenfigureheight,keepaspectratio]{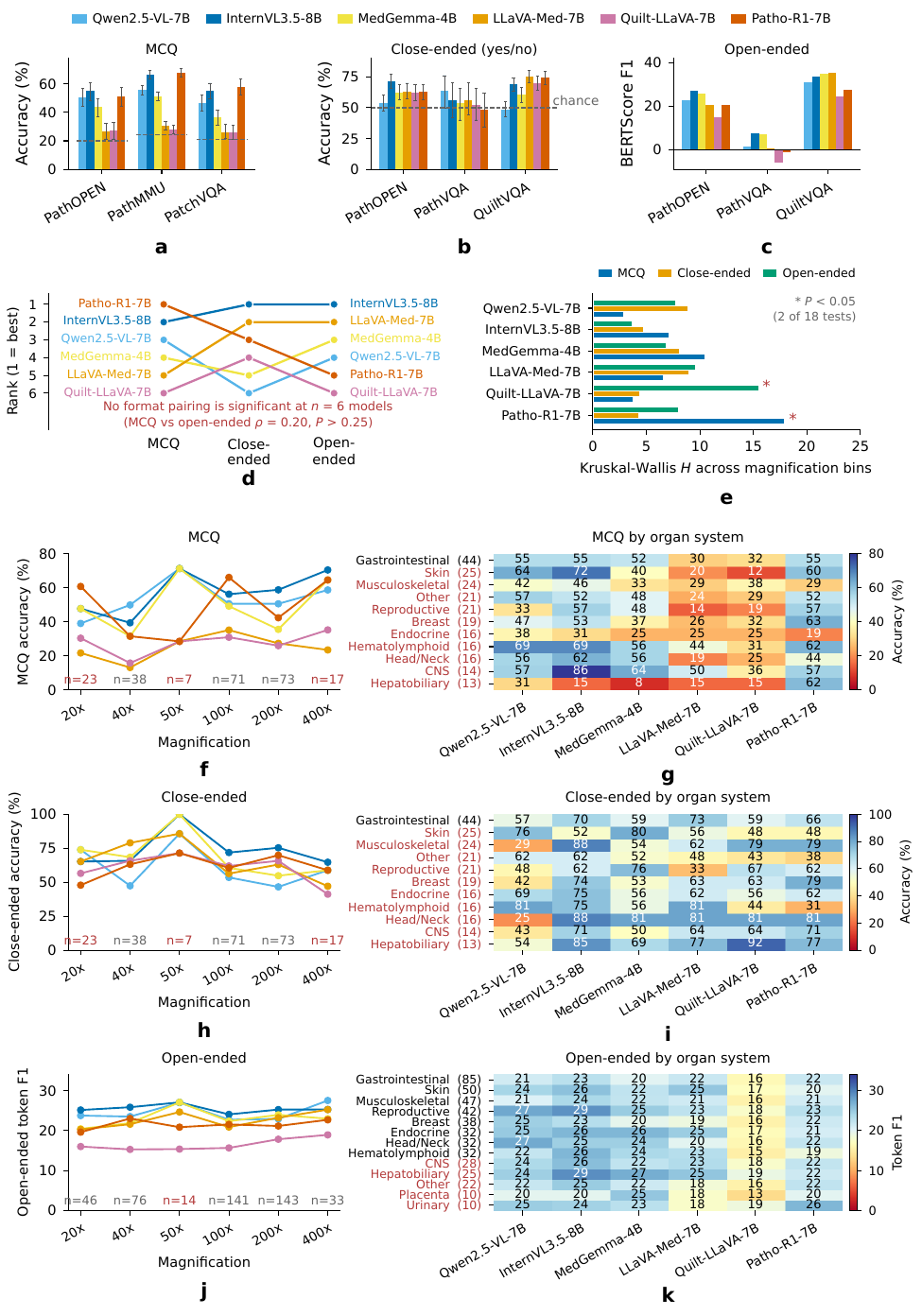}}{\textbf{Zero-shot answering across datasets, formats and metadata strata.}
\textbf{a}--\textbf{c}, Per-model MCQ accuracy, close-ended accuracy and open-ended
BERTScore F1. Dashed lines mark dataset-specific MCQ chance
levels (2--16 options) and the 50\% close-ended chance level. BERTScore F1 is rescaled
so that zero denotes similarity no greater than unrelated text; filtered PathVQA falls
below zero for two of six models. \textbf{d}, Model ranks within each format (1 = best).
Table~\ref{tab:format-ranks} gives the Spearman rank correlations.
\textbf{e}, Kruskal--Wallis $H$ for performance by magnification, for each model and
format; 2 of 18 tests reach uncorrected $P < 0.05$.
\textbf{f}--\textbf{k}, Performance by magnification (left) and organ system (right)
for MCQ (\textbf{f}, \textbf{g}), close-ended (\textbf{h}, \textbf{i}) and open-ended
questions (\textbf{j}, \textbf{k}). Item counts are printed per bin, with counts below
$n = 30$ in red. The comparator records used here lack this metadata.
\textbf{a} and \textbf{b} show bootstrap 95\% confidence intervals; \textbf{c} shows
point estimates. Open-ended results in \textbf{d}, \textbf{e}, \textbf{j} and \textbf{k}
use token F1 (Section~\ref{subsubsec:zeroshot}). Section~\ref{sec:pillar4} gives the
rank and magnification tests and their interpretation. Table~\ref{tab:dataset-comparison}
reports pooled comparisons.}{fig:pillar4}

\PathOPENFigure{\includegraphics[width=\linewidth,height=\pathopenfigureheight,keepaspectratio]{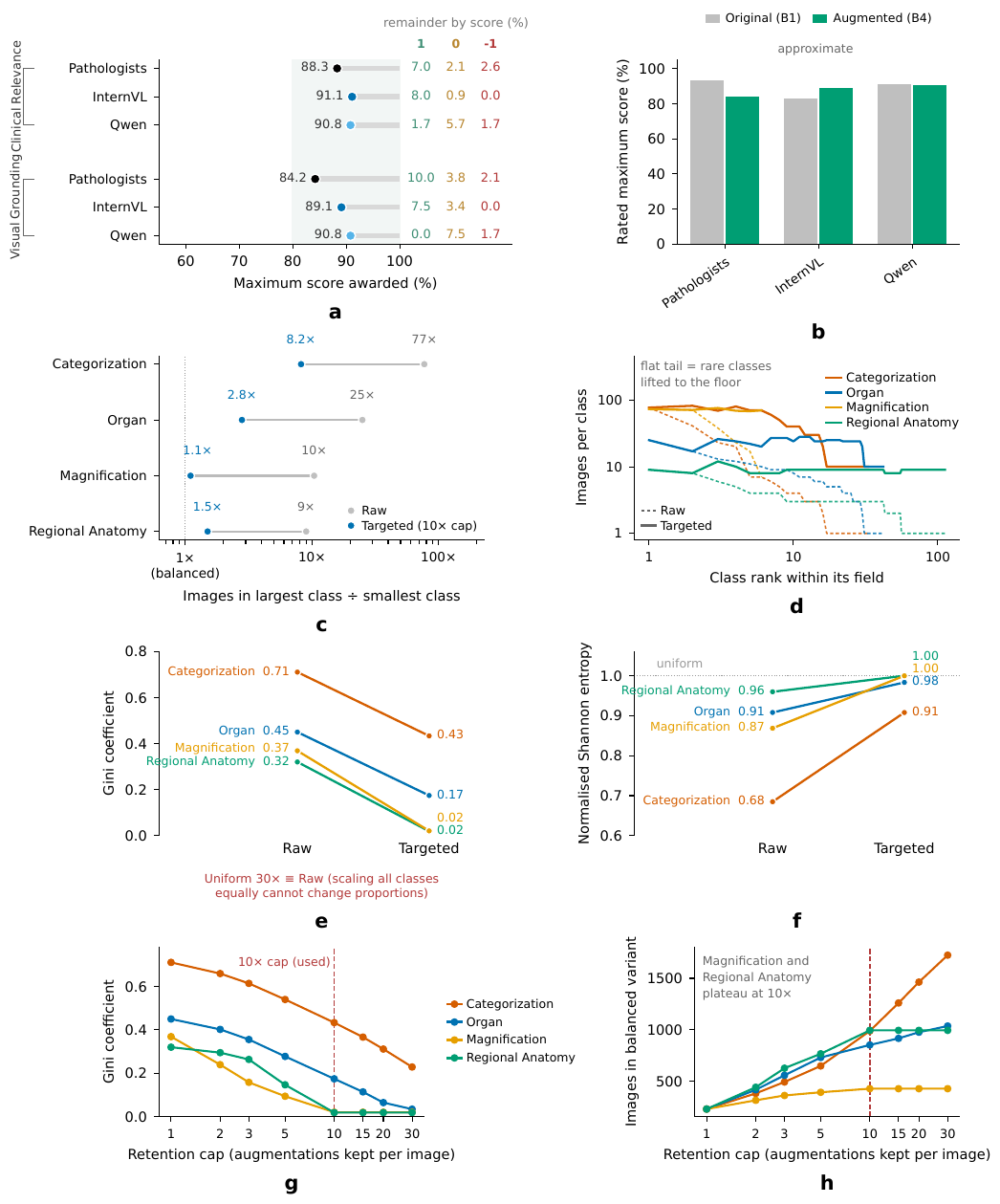}}{\textbf{Ratings of augmented images and class balance after selecting subsets.}
\textbf{a}, Benchmark~4 ratings from five pathologists and two judges. Dots show the
percentage receiving the maximum score; grey bars show the remaining percentage,
broken down by score at right. \textbf{b}, Visual Grounding ratings on augmented
images (Benchmark~4) and originals (Benchmark~1), using unpaired pools and differently
worded criteria. \textbf{c}, Largest-to-smallest class ratios before and after
field-specific retention, on a logarithmic scale. \textbf{d}, Ranked class counts
before retention (dashed) and after retention (solid), including the flat tail of
retained rare classes. \textbf{e}, \textbf{f}, Gini coefficients and normalised Shannon
entropy. Uniform 30$\times$ augmentation has the same class proportions as the originals
and is omitted from these panels. \textbf{g}, \textbf{h}, Gini coefficients and selected
image counts across the retention-cap sweep (Section~\ref{subsubsec:balancing}).
Colours identify metadata fields. The four manifests each select existing augmentation
files for one field and may overlap. Section~\ref{sec:pillar5} gives the balance results;
Table~\ref{tab:benchmark4-rating} reports augmented-image ratings and agreement.}{fig:pillar5}

\FloatBarrier
\section{Usage Notes}\label{sec:usage-notes}

\subsection{Access and licence}

Data availability identifies the dataset version and licence. Code availability
links to the analysis scripts. Section~\ref{subsec:file-records} describes the files.

\subsection{Loading records and handling missing data}\label{subsec:usage-loading}

For the Zenodo release, read \texttt{data.json} from the package root. On Hugging
Face, \texttt{xuzhl/PathOPEN} provides the \texttt{cases} and \texttt{originals}
configurations. Pair each non-empty question with
every available original image in its case and keep the \texttt{Case ID} with
each pair. Skip empty image slots and keep open-ended questions that have only a
correct answer. Count each question once per case for question totals and once
per available image for the corresponding pair counts in
Table~\ref{tab:data-records}.

Resolve filenames in \texttt{Original} and \texttt{Augmented} using the same
\texttt{Directory}, relative to the package root. The package \texttt{README.md}
explains how to load the data and run \texttt{load\_pathopen.py} to verify record
links and file checksums.

\paragraph{Normalising metadata.}
Organ, categorization and regional anatomy retain their collected spelling and case.
For analyses that need canonical labels, define and document a mapping from the
stored values. Section~\ref{subsec:metadata-records} gives the counts after merging
capitalisation variants. The package uses \texttt{Gastrointestinal - Small Intestine}.
Allow for the earlier spelling when linking historical records.
Supplementary Table~\ref{tab:label-abbreviations} explains the abbreviations used
in figure axes.

\subsection{Grouping records before splitting}\label{subsec:usage-splits}

Assign splits using \texttt{Case ID} before selecting images, questions or augmentations.
Keep each case's images, questions and augmentations in the same split. If
patient linkage is available, group cases from the same patient together. A test
augmentation can share its answer with the original image or another augmentation
in training. Treat derived views of a case as repeated observations.

\subsection{Constructing tasks and reporting scores}

\paragraph{Question answering.}
Report scores separately for the three question formats.
The six models ranked differently depending on the format
(Section~\ref{sec:pillar4}). For MCQs, pass all five options to the model when
measuring accuracy. For cross-dataset comparisons, account for differences in
option counts by reporting accuracy above chance. PathOPEN's chance level
is 20\%. Section~\ref{subsubsec:zeroshot} gives the answer parsing and metric conventions,
including counting unparseable MCQ responses as incorrect and the rescaled BERTScore
baseline for open-ended comparisons.

\paragraph{Stratified analysis and metadata use.}
Report how groups are defined, how many items they contain and how repeated
cases enter the analysis. Most organ groups have fewer than 30 MCQ items.
Magnification counts also vary by format (Section~\ref{sec:pillar4}). The metadata
can be used as retrieval context or to select examples for a training task.

\paragraph{Preference pairs and rubric scores.}
Pair the correct and wrong answers for the same question, case and image
to construct preference examples. The open-ended records provide such pairs for
201 of 309 questions. An MCQ gives four pairs by matching each distractor
with the correct option. The same pathologist wrote the two answers in a pair.
Readers with the historical Benchmark~2 scores can also select pairs by
pathologist-assigned difficulty (Section~\ref{sec:pillar2}). To link a score
to a released answer, match the case, image, question and answer text. Of the 916 available-image MCQ
distractors, 912 have two historical scores by image and option position. Two of
those distractors in case 36 have different text in the historical spreadsheet,
leaving 910 with matching text and two recorded scores.

Keep rubric scores distinct from answering-model accuracy. On Benchmark~2, a higher
score describes a subtler wrong answer; it does not indicate that the answer is correct.
A score of $-1$ records inability to evaluate and carries no difficulty tier. A blank
score cell indicates a missing rating. For summary statistics, follow the
analysis-specific $-1$ conventions in Section~\ref{subsubsec:stats}.
For Benchmark~5b comparisons, report both judges
because their dataset rankings reversed
(Section~\ref{sec:pillar3}); Benchmark~5 has no human baseline.

\subsection{Using augmented images and balanced manifests}

Choose a manifest for the field you want to balance. Resolve
\texttt{relative\_path} against the package root and use the source case and
image identifiers to keep the selected file in its case's split. The manifests
can share files. Since regional anatomy is nested within organ, a selection
by either field may also change the distribution of the other. Section~\ref{subsubsec:balancing}
gives the selection rule and seed.

\subsection{Scope of reuse}

PathOPEN contains 157 cases from one institution, collected over three months by two
pathologists. Selection favoured interesting teaching cases and diagnoses encountered
during that window, with balancing between normal and abnormal cases.
Case-level metadata Gini coefficients range from 0.24 to 0.66, so rare-class strata
contain few observations (Figure~\ref{fig:composition}g).

Each human-rated item has one rating, so differences among pathologists reflect
their separate item pools. Since most scores were at the maximum, interpret
$\kappa$ alongside exact agreement and AC1 (Section~\ref{sec:pillar1}).

\FloatBarrier

\section*{Data availability}\label{sec:data_availability}
PathOPEN v1.0.0 is available on Zenodo~\citep{naeem2026pathopen} under CC BY 4.0.
The version DOI is \href{https://doi.org/10.5281/zenodo.24358341}{%
  \nolinkurl{10.5281/zenodo.24358341}}.
The archive contains 157 case records, 229 original images, 6,870 augmented images
and four manifests. \href{https://huggingface.co/datasets/xuzhl/PathOPEN}{Hugging Face}
provides the case records and original images as Parquet files.
The complete historical item-level pathologist and VLM rating tables for
Benchmarks~1--3 are held separately from these releases, as are the 174 images
and item-level ratings used for Benchmark~4 (Section~\ref{subsubsec4_5_3}).
See Sections~\ref{sec:data-records} and \ref{sec:usage-notes} for file layout and loading.
Obtain comparator datasets from their original sources and see
Section~\ref{subsec:pathvqa-selection} for the PathVQA filter.

\section*{Code availability}\label{sec:code_availability}
The \href{https://github.com/embedded-robotics/path-open-data}{PathOPEN code repository}
contains scripts for processing the data, augmenting images, building manifests,
evaluating models and generating the figures and tables. The
\href{https://github.com/embedded-robotics/path-open-data/tree/4660b400d633e9ad8f875c1f0985f0eed6c8d787}{v1.0.0 code snapshot}
corresponds to the Zenodo release.

\clearpage
\setcounter{table}{0}
\renewcommand{\thetable}{S\arabic{table}}
\setcounter{figure}{0}
\renewcommand{\thefigure}{S\arabic{figure}}
\renewcommand{\theHfigure}{supplement.\arabic{figure}}

\section*{Supplementary Information}
\label{sec:supplementary}

Supplementary Figure~\ref{fig:pathopen-text} shows PathOPEN word counts and
question openings. Supplementary Note~1 explains how they were counted.
\hyperref[note:metadata]{Supplementary Note~2} describes metadata coverage.
Supplementary Tables~\ref{tab:benchmark1}--\ref{tab:benchmark4} give the rubrics
used by pathologists and VLM judges. Only the judges used the rubrics in
Supplementary Tables~\ref{tab:benchmark5a} and \ref{tab:benchmark5b}.
Each rubric uses $\{-1,0,1,2\}$. The rater assigns $-1$ when an item cannot be
evaluated (Section~\ref{subsubsec:stats}). Supplementary
Table~\ref{tab:label-abbreviations} gives the axis-label abbreviations.

\subsection*{Supplementary Note 1. Text summaries and comparator profiles}

Supplementary Figure~\ref{fig:pathopen-text} shows PathOPEN word counts and question
openings. We report the percentages with at least five and at least ten words.
Strict thresholds would change the percentages by two to five points. The
open-ended summaries include every non-missing entry across both question slots.
For MCQ and close-ended fields, we counted duplicate wording once per column.
This excludes two MCQ questions, three close-ended questions and ten options
from these summaries. The dataset retains all these records.

Almost all native open-ended questions and correct and wrong answers have at least
5 words. At least 10 words occur in 86.4\% of questions, 98.4\% of correct answers and
99.0\% of wrong answers. The rounded mean lengths are 18 words for questions, 23 for correct answers
and 21 for wrong answers (Supplementary Figure~\ref{fig:pathopen-text}a). All MCQ questions and 65\% of options have at least 5 words;
95\% of questions and 48\% of options have at least 10 words
(Supplementary Figure~\ref{fig:pathopen-text}b,e). MCQ questions average 19 words on the same rounded scale. In the option-length
summary, options average 8.6 words (roughly 9), with median 9; 58.6\% exceed 6 words and
10.6\% fall within the requested 4--6-word band. These option statistics use the per-column distinct
strings described above. Counting options per question--image record, as in Benchmark~5, gives a mean of
9.67 words. Pathologists attributed the shorter options, relative to native open-ended
answers, to keeping all choices relevant to the question and correct choice without
unnecessary detail. Close-ended questions average about 12 words
(Supplementary Figure~\ref{fig:pathopen-text}c). Of the 154 distinct close-ended
questions, 152 (98.7\%) contain at least five words and 101 (65.6\%) contain at
least ten. Panel \textbf{e} shows these percentages rounded to whole numbers.

The open-ended questions begin with 18 distinct words: \emph{what} accounts for 65\%,
\emph{how} for 9\%, \emph{considering} for 7\%, \emph{in} for 6\%, and \emph{assuming}
for 3\%; the remaining 13 openings account for 9\%
(Supplementary Figure~\ref{fig:pathopen-text}d). Openings such as \emph{considering}, \emph{in} and
\emph{assuming} introduce contextual premises. The common close-ended openings are
\emph{is} (40\%), \emph{does} (25\%) and \emph{would} (10\%).
These summaries describe the construction of the questions; the pathologist-rated
assessment is reported in Section~\ref{sec:pillar1}.

\paragraph{PathVQA comparator profile.}
The comparator datasets are distributed by their authors and are not redistributed
with PathOPEN. Section~\ref{subsec:pathvqa-selection} specifies the two-stage PathVQA
filter. The image-selection stage retains 24.8\% of the 4,998 raw images and 27.4\% of
the question--answer pairs (Figure~\ref{fig:pathvqa-selection-text}a).

Figure~\ref{fig:pathvqa-selection-text}b,c compares raw PathVQA with the
image-filtered set before filtering for length. At least five words occur in
45.5\% of raw open-ended questions and 50\% after image filtering. At ten words,
the shares are 9\% and 13\%, respectively. For answers, 15\% reach five words in
both sets, and 4\% and 3\% reach ten. The distributions are right-skewed. Raw
questions average 5.3 words (median four) and raw answers average 2.6 (median
one). After image filtering, the rounded means are six words for questions and three
for answers (Figure~\ref{fig:pathvqa-selection-text}b). These short references matter for cross-dataset scoring
(Section~\ref{sec:pillar4}).

\subsection*{Supplementary Note 2. Metadata coverage}\label{note:metadata}

At case level, Gini coefficients are 0.44 for organ, 0.66 for categorization, 0.24 for
regional anatomy and 0.37 for magnification (Figure~\ref{fig:composition}g).
Image-weighted values are 0.45, 0.71, 0.32 and 0.37 respectively, as used for augmentation
balancing in Section~\ref{sec:pillar5}. Categorization is the most concentrated field
and regional anatomy the least under both weightings.

Of the cases with available images, 118 have one magnification and 37 have
more than one. The most frequent combinations are 100$\times$ with 200$\times$
(24 cases) and 40$\times$ with 100$\times$ (22 cases). A case with three
magnifications contributes to three bins (Figure~\ref{fig:composition}f).
Figure~\ref{fig:composition}h crosses the 13 most frequent organ systems with
the eight most frequent categories, covering 85\% of cases. The data cover 41
of the 104 possible combinations. Gastrointestinal infection is the largest
group, with 12 cases.

\PathOPENFigure{\includegraphics[width=\linewidth,height=\pathopenfigureheight,keepaspectratio]{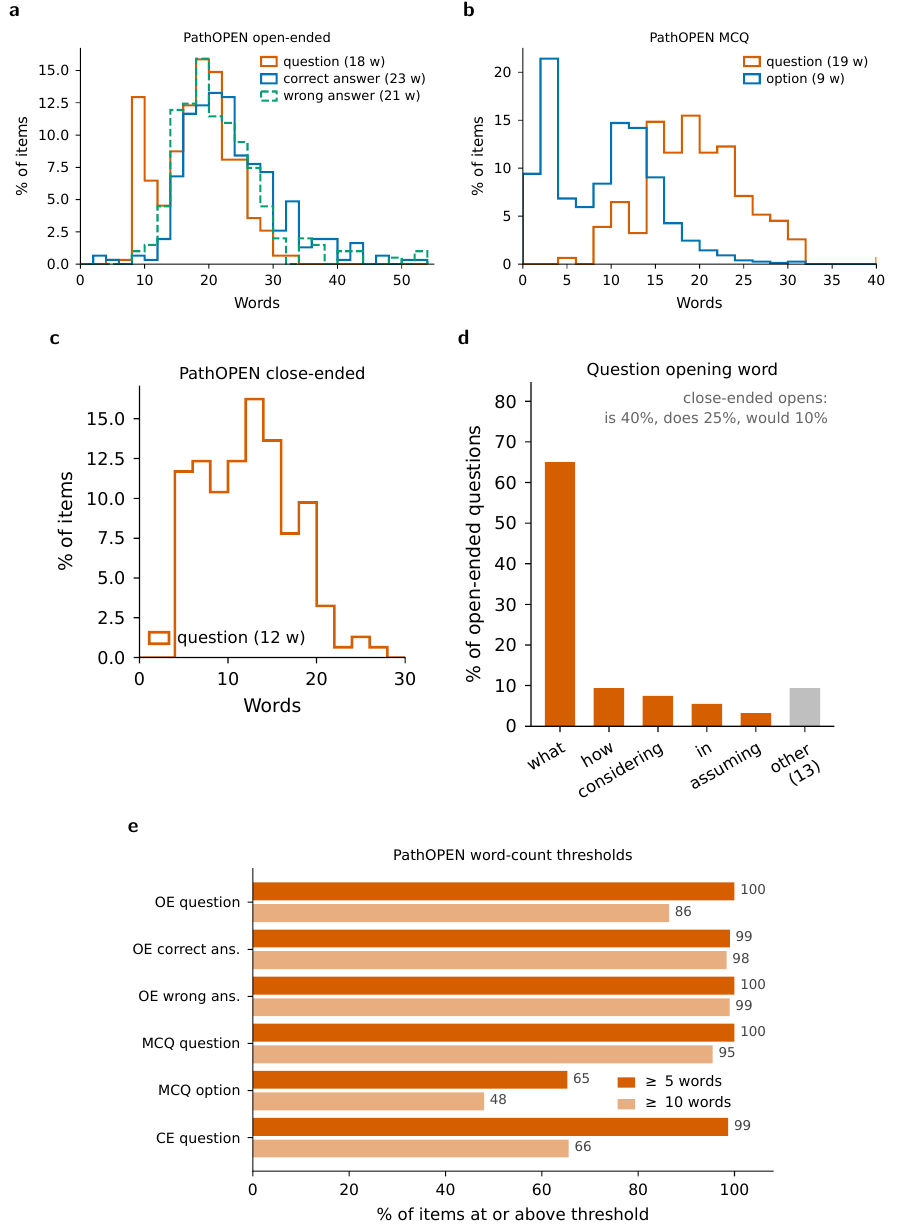}}{\textbf{Text characteristics of PathOPEN questions and answers.}
\textbf{a}--\textbf{c}, Word-count distributions for open-ended questions and answers,
multiple-choice questions (MCQs) and options, and close-ended questions, respectively.
\textbf{d}, The five most frequent opening words of open-ended questions and the pooled
remainder, with the three most frequent close-ended openings annotated.
\textbf{e}, Percentages of entries with at least five and at least ten words.
Open-ended summaries include all non-missing entries. MCQ and close-ended summaries
count distinct strings within each column.}{fig:pathopen-text}

\begin{table}[!ht]
    \centering
    \small
    \caption{Benchmark 1 scoring rubric for correct open-ended answers.}\label{tab:benchmark1}
    \vspace{1.5em} 

    \begin{tabularx}{\linewidth}{@{}lX@{}}
        \toprule
        \multicolumn{2}{c}{\textbf{Knowledge Interpretation and Deduction}} \\
        \midrule
        \textbf{Score} & \textbf{Reasoning} \\
        \midrule
        $-1$ & Unable to comprehend the question and/or image, or unable to make the evaluation \\
        0 & Answer did not address the question correctly, or only addressed minor/unrelated parts of the question, or did not use correct deductive reasoning when provided \\
        1 & Answer partially addresses the question by leaving out some important details and/or use partially correct deductions \\
        2 & Answer addresses the majority details about the question correctly without the inclusion of any unrelated suggestions and/or provides reasonable deductions to achieve the answer \\
        \bottomrule
    \end{tabularx}

    \vspace{1em} 

    \begin{tabularx}{\linewidth}{@{}lX@{}}
        \multicolumn{2}{c}{\textbf{Visual Grounding}} \\
        \midrule
        \textbf{Score} & \textbf{Reasoning} \\
        \midrule
        $-1$ & Unable to comprehend the question and/or image, or unable to make the evaluation \\
        0 & Answer appears to focus on context clues from text of the question without considering the image information at all or considers the inconsequential parts of the image having no connection to question \\
        1 & Answer only tangentially considers the image related to question, but still appears to rely on textual context \\
        2 & Answer identified and incorporated the image to directly answer the question \\
        \bottomrule
    \end{tabularx}
\end{table}

\begin{table}[!ht]
    \centering
    \small
    \caption{Benchmark 2 scoring rubric for wrong open-ended answers.}\label{tab:benchmark2}
    \vspace{1em} 

    \begin{tabularx}{\linewidth}{@{}lX@{}}
        \toprule
        \multicolumn{2}{c}{\textbf{Error Proximity and Deductive Plausibility}} \\
        \midrule
        \textbf{Score} & \textbf{Reasoning} \\
        \midrule
        $-1$ & Unable to comprehend the question and/or image, or unable to make the evaluation \\
        0 & Answer reflects a wrong conceptual frame or unrelated inference, contradicts key constraints, or ignores the question’s core; reasoning is largely irrelevant or speculative. \\
        1 & Answer shows partial task understanding with mixed correct/incorrect deductions, but misses one or more critical constraints leading to a materially wrong conclusion. \\
        2 & Answer is a near-miss: reasoning path is appropriate and mostly correct, with a single fine-grained mistake or omission causing a subtle deviation from the correct answer. \\
        \bottomrule
    \end{tabularx}

    \vspace{1em} 

    \begin{tabularx}{\linewidth}{@{}lX@{}}
        \multicolumn{2}{c}{\textbf{Visual Grounding Error}} \\
        \midrule
        \textbf{Score} & \textbf{Reasoning} \\
        \midrule
        $-1$ & Unable to comprehend the question and/or image, or unable to make the evaluation \\
        0 & Answer is not grounded in the image (relies on textual cues or irrelevant regions) or cites visual elements not present. \\
        1 & Answer references some relevant visual content but mislocalizes or misinterprets key cues (e.g., confuses similar objects/regions), leading to a noticeable error. \\
        2 & Answer attends to the correct region and most pertinent features; the mistake arises from a fine-grained visual ambiguity (e.g., off-by-one count, subtle attribute/relationship) \\
        \bottomrule
    \end{tabularx}
\end{table}

\begin{table}[!ht]
    \centering
    \small
    \caption{Benchmark 3 scoring rubric for close-ended answers.}\label{tab:benchmark3}
    \vspace{1em} 
    \begin{tabularx}{\linewidth}{@{}lX@{}}
        \toprule
        \multicolumn{2}{c}{\textbf{Visual Grounding/Reasoning}} \\
        \midrule
        \textbf{Score} & \textbf{Reasoning} \\
        \midrule
        $-1$ & Unable to comprehend the question and/or image, or unable to make the evaluation \\
        0 & Question can be answered by focusing solely on context clues from text without considering the image information at all \\
        1 & Question can be answered by tangentially considering the image information, but still the textual context plays a major role \\
        2 & Question can be answered by carefully combining contextual clues from text and visual clues from the image \\
        \bottomrule
    \end{tabularx}
\end{table}

\begin{table}[!ht]
    \centering
    \small
    \caption{Benchmark 4 scoring rubric for augmented images.}\label{tab:benchmark4}
    \vspace{1em} 

    \begin{tabularx}{\linewidth}{@{}lX@{}}
        \toprule
        \multicolumn{2}{c}{\textbf{Clinical Relevance}} \\
        \midrule
        \textbf{Score} & \textbf{Reasoning} \\
        \midrule
        $-1$ & Unable to comprehend the question and/or image, or unable to make the evaluation \\
        0 & Image has minimal clinical relevance which makes it highly unlikely to form any sort of diagnosis \\
        1 & Image is clinically relevant to form majority of diagnosis but some patches can mislead the user into creating artificial diagnosis \\
        2 & Image is clinically relevant to form majority of diagnoses correctly without any misleading patches \\
        \bottomrule
    \end{tabularx}

    \vspace{1em} 

    \begin{tabularx}{\linewidth}{@{}lX@{}}
        \multicolumn{2}{c}{\textbf{Visual Grounding}} \\
        \midrule
        \textbf{Score} & \textbf{Reasoning} \\
        \midrule
        $-1$ & Unable to comprehend the question and/or image, or unable to make the evaluation \\
        0 & Image contains misleading information or does NOT contain enough information to be combined with question text to form the diagnosis given in the answer \\
        1 & Image contains appropriate information which complements the question text to form majority of the diagnosis given in the answer, but the answer contains some information which cannot be linked back to image because of corrupt or misleading image patches \\
        2 & Image contains appropriate information which along with the question text can be used to form complete diagnosis given in answer without any information NOT linking back to image \\
        \bottomrule
    \end{tabularx}

\end{table}

\begin{table}[!ht]
    \centering
    \small
    \caption{Benchmark 5a scoring rubric for the correct multiple-choice question (MCQ) option.}\label{tab:benchmark5a}
    \vspace{1em} 

    \begin{tabularx}{\linewidth}{@{}lX@{}}
        \toprule
        \multicolumn{2}{c}{\textbf{Standalone Completeness}} \\
        \midrule
        \textbf{Score} & \textbf{Reasoning} \\
        \midrule
        $-1$ & Unable to comprehend the question, option, and/or image, or unable to make the evaluation \\
        0 & Option is a bare label, category name, or fragment (e.g., single word/short phrase) that only makes sense when read alongside the other options; conveys no independent clinical claim \\
        1 & Option conveys a partial claim but relies on implicit reference to the question stem or option list to be understood as a complete statement \\
        2 & Option reads as a complete, standalone clinical statement that would remain fully meaningful if presented without the question's MCQ framing or the other options \\
        \bottomrule
    \end{tabularx}

    \vspace{1em} 

    \begin{tabularx}{\linewidth}{@{}lX@{}}
        \multicolumn{2}{c}{\textbf{Visual Grounding}} \\
        \midrule
        \textbf{Score} & \textbf{Reasoning} \\
        \midrule
        $-1$ & Unable to comprehend the question and/or image, or unable to make the evaluation \\
        0 & Option can be identified as correct using only the question text and/or general medical knowledge, without needing to examine the image \\
        1 & Option is only tangentially supported by the image; the question text or option phrasing provides most of the disambiguating information \\
        2 & Option can only be confirmed as correct by directly examining and interpreting relevant visual features in the image \\
        \bottomrule
    \end{tabularx}
\end{table}

\begin{table}[!ht]
    \centering
    \small
    \caption{Benchmark 5b scoring rubric for multiple-choice question (MCQ) distractors.}\label{tab:benchmark5b}
    \vspace{1em} 

    \begin{tabularx}{\linewidth}{@{}lX@{}}
        \toprule
        \multicolumn{2}{c}{\textbf{Distractor Standalone Plausibility}} \\
        \midrule
        \textbf{Score} & \textbf{Reasoning} \\
        \midrule
        $-1$ & Unable to comprehend the question, option, and/or image, or unable to make the evaluation \\
        0 & Option is an implausible or nonsensical standalone claim, easily dismissed without reference to the image (e.g., wrong organ system, internally contradictory, or too vague to evaluate) \\
        1 & Option is a plausible standalone claim in general clinical terms but bears little morphological or contextual resemblance to the correct option for this specific case \\
        2 & Option is a plausible, well-formed standalone claim that closely parallels the correct option in structure and clinical register, differing only in a specific factual, diagnostic, or morphological detail \\
        \bottomrule
    \end{tabularx}

    \vspace{1em} 

    \begin{tabularx}{\linewidth}{@{}lX@{}}
        \multicolumn{2}{c}{\textbf{Distractor Visual Grounding Error}} \\
        \midrule
        \textbf{Score} & \textbf{Reasoning} \\
        \midrule
        $-1$ & Unable to comprehend the question and/or image, or unable to make the evaluation \\
        0 & Option shows no reference to plausible visual features of the image; wrong purely on textual/conceptual grounds \\
        1 & Option references visual features that are present in the image but misattributes or misinterprets them (e.g., correct structure, wrong finding) \\
        2 & Option references visual features precisely enough that distinguishing it from the correct option requires fine-grained visual discrimination, not just textual knowledge \\
        \bottomrule
    \end{tabularx}
\end{table}

\begin{table}[t]
\centering\small
\caption{\textbf{Axis-label abbreviations for the metadata figures.}
Stored values retain their full spelling and case. These abbreviated display labels
are used on figure axes, including Figure~\ref{fig:composition}, and do not replace
the stored metadata values.}
\label{tab:label-abbreviations}
\begin{tabular}{l l}
\toprule
Stored value & Axis label \\
\midrule
\multicolumn{2}{l}{\emph{Organ-system prefix}} \\
Gastrointestinal & GI \\
Musculoskeletal & MSK \\
Hematolymphoid & Heme \\
Central Nervous System & CNS \\
Hepatobiliary & Hepatobil. \\
Cardiovascular & CV \\
Reproductive & Reprod. \\
Respiratory & Resp. \\
\addlinespace
\multicolumn{2}{l}{\emph{Other long terms}} \\
Small Intestine & Small Intest. \\
(Tissue degeneration) & (degen.) \\
(autoimmune) & (AI) \\
(Overgrowth) & (overgr.) \\
Intestinal villous mucosa & Intestinal villous muc. \\
Subcutaneous tissues & Subcutaneous tiss. \\
Squamous epithelium & Squamous epith. \\
\bottomrule
\end{tabular}
\end{table}

\FloatBarrier

\clearpage
\begin{appendices}

\section{Pathology categorization labels}\label{sec:appendix_categories}
The table lists selected categories and types for \texttt{Categorization}.
The dataset stores \texttt{Injury (Bleeding} with an unmatched parenthesis
and \texttt{Injury (Tissue degeneration)} as written. Use the stored strings
when matching records.

\vspace{1em}
\noindent\begin{tabularx}{\textwidth}{@{}>{\normalsize}l>{\normalsize\itshape}X@{}}
    \toprule
    \textbf{Category} & \textbf{Types} \\
    \midrule
    Normal & Normal \\
    \midrule
    Infection & Benign or unspecified \\
    \midrule
    Injury & Bleeding, Tissue degeneration \\
    \midrule
    Neoplasia & Benign, Malignant \\
    \midrule
    Hyperplasia & Overgrowth, Not Neoplastic \\
    \midrule
    Other & Free Text \\
    \bottomrule
\end{tabularx}

\clearpage
\section{Organ labels and subdivisions}\label{sec:appendix_organs}

The table lists organ labels and subdivisions used in the collection vocabulary.
Use \texttt{data.json} for exact stored values. The organ label
\texttt{Gastrointestinal - Dudenum} retains the original spelling. The regional
anatomy field spells it \texttt{Duodenum}.

\vspace{1em}
\noindent\begin{tabularx}{\textwidth}{@{}>{\normalsize}l>{\normalsize\itshape}X@{}}
    \toprule
    \textbf{Organ} & \textbf{Subdivisions} \\
    \midrule
    Breast & Breast Ducts, Breast Glands \\
    \midrule
    Cardiovascular & Blood Vessels, Heart, Lymphatics, Veins \\
    \midrule
    Central Nervous System & Cerebellum, Cerebrum, Spinal Cord, Ventricles \\
    \midrule
    Endocrine & Adrenal Gland, Pancreas, Parathyroid, Pineal Gland, Pituitary Gland, Thyroid, Salivary Glands \\
    \midrule
    Eye & Eye, Optic Nerve \\
    \midrule
    Gastrointestinal & Colon, Dudenum, Esophagus, Rectum, Small Intestine, Stomach \\
    \midrule
    Head/Neck & Ear, Mouth, Nose, Throat \\
    \midrule
    Hematolymphoid & Lymph Nodes, Bone Marrow, Spleen, Thymus \\
    \midrule
    Hepatobiliary & Gall Bladder, Liver \\
    \midrule
    Musculoskeletal & Bone, Soft Tissue \\
    \midrule
    Reproductive & Cervix, Fallopian Tube, Ovary, Penis, Prostate, Testis, Uterus, Vagina \\
    \midrule
    Respiratory & Lung, Trachea \\
    \midrule
    Skin & Skin \\
    \midrule
    Urinary & Kidney, Bladder \\
    \midrule
    Other & Free Text \\
    \bottomrule
\end{tabularx}

\end{appendices}

\end{document}